\documentclass[lettersize,journal]{IEEEtran}
\usepackage{amsmath,amsfonts}
\usepackage{algorithmic}
\usepackage{algorithm}
\usepackage{array}
\usepackage[caption=false,font=normalsize,labelfont=sf,textfont=sf]{subfig}
\usepackage{textcomp}
\usepackage{stfloats}
\usepackage{url}
\usepackage{verbatim}
\usepackage{graphicx}
\usepackage{cite}
\usepackage{amsmath}
\usepackage{mathtools}
\usepackage{tabularx}
\usepackage{tabularray}
\usepackage{longtable}
\usepackage{xtab}
\usepackage{float}

\usepackage[english]{babel}
\usepackage[dvipsnames,svgnames,x11names]{xcolor}

\def\BibTeX{{\rm B\kern-.05em{\sc i\kern-.025em b}\kern-.08em
    T\kern-.1667em\lower.7ex\hbox{E}\kern-.125emX}}

\begin{document}
\newlength{\saveparindent}
\newcommand{\indentpar}{\par\hspace*{\saveparindent}\ignorespaces}

\title{Emergency Collision Avoidance and Mitigation
Using Model Predictive Control and Artificial Potential Function\\
}

\author{Xu Shang,~\IEEEmembership{Member,~IEEE,}
        Azim Eskandarian,~\IEEEmembership{Senior Member,~IEEE}
\thanks{Manuscript submitted 11 December 2022; Revised 19 January 2023; Accepted 26 January 2023 }
\thanks{The authors are with the Autonomous Systems and Intelligent Machines Laboratory, Virginia Tech, Blacksburg, VA 24061, USA.(e-mail: shangxu@vt.edu; eskandarian@vt.edu). }
}


\maketitle

\begin{abstract}
Although extensive research in emergency collision avoidance has been carried out for straight or curved roads in a highway scenario, a general method that could be implemented for all road environments has not been thoroughly explored. Moreover, most current algorithms don't consider collision mitigation in an emergency. This functionality is essential since the problem may have no feasible solution. We propose a safe controller using model predictive control and artificial potential function to address these problems. A new artificial potential function inspired by line charge is proposed as the cost function for our model predictive controller. The vehicle dynamics and actuator limitations are set as constraints. The new artificial potential function considers the shape of all objects. In particular, the artificial potential function we proposed has the flexibility to fit the shape of the road structures, such as the intersection. We could also realize collision mitigation for a specific part of the vehicle by increasing the charge quantity at the corresponding place. We have tested our methods in 192 cases from 8 different scenarios in simulation with two different models. The simulation results show that the success rate of the proposed safe controller is 20$\%$ higher than using HJ-reachability with system decomposition by using a unicycle model. It could also decrease 43$\%$ of collision that happens at the pre-assigned part. The method is further validated in a dynamic bicycle model.  
\end{abstract}


\section{Introduction}
\subsection{Background}
\begin{figure}[!ht]
\centering
\includegraphics[width=0.5\textwidth]{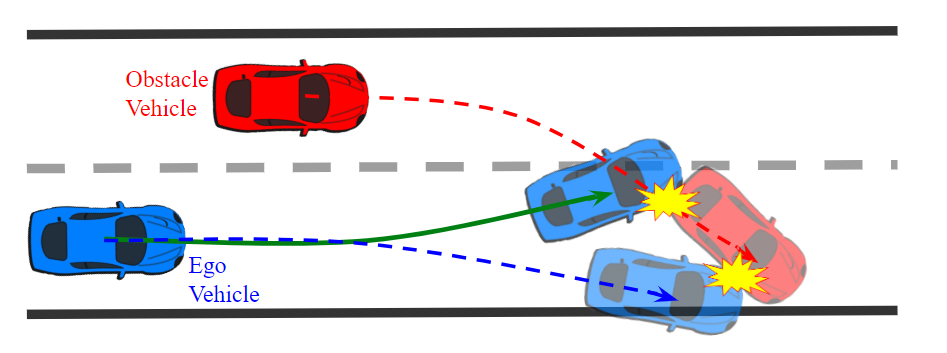}
\caption{Strategy Illustration}
\label{SI}
\end{figure}

Almost 3700 people are killed globally every day because of collisions on roads \cite{1CDC}. These collisions could be caused by driver errors, drowsy driving, or driving under influence(DUI). On the other hand, a malfunction in the car, such as a flat tire or brake failure, could also result in collisions. Autonomous driving has been considered as a key to solve these problems \cite{3EA}. It will not have problems of losing concentration like humans and could behave better when other vehicles malfunction and cause an emergency because of its powerful computational speed. 
\indentpar
Autonomous driving has been studied for around 40 years and more and more research effect keep devoting to it recently with the development of computers and sensors. Multiple technologies related to autonomous driving have been already implemented to help people in daily driving such as lane keeping, adaptive cruise control, and automatic parking \cite{4EA}. $100\%$ autonomous driving is expected to be realized in 5-10 years which could significantly improve traffic safety and protect people's lives. However, in some emergencies in which vehicles move at high speeds and there exists less space for maneuvering, even $100\%$ autonomous driving may fail to find a path due to few feasible solutions or even no feasible solution. In this case, collision mitigation becomes extremely important but it lacks sufficing investigations. 
\indentpar
Based on the fact that the average occupancy of a vehicle is 1.5 people in 2018 ~\cite{5Davis}, our strategy for collision mitigation is to protect the place occupied by passengers and try to  redirect the collision to happening at the vehicle's body without passengers. For instance, an inevitable collision happens in Fig. \ref{SI}. The red obstacle vehicle loses control at high speed and turns to the right lane. The blue ego vehicle could either keep turning right as the blue dash line or turn left to change the lane as the green dash line. Assuming only a driver is in the blue car, our method will tend to choose the green dash line to protect the driver.  
\vspace{-0.4cm}

\indentpar
This research aims to develop a safe controller for emergency collision avoidance for all road structures. The safe controller will try to find a safe path or protect the human-occupied part of the vehicle if the collision is inevitable. In the rest of this paper, the ego vehicle represents the autonomous vehicle that we could control, while the obstacle vehicle is the vehicle that behaves abnormally and causes a hazard or potential harm.

\subsection{Related Work}
Emergency collision avoidance is a subproblem of path planning. Compared with normal path planning problems, the environment is more complicated and has less time to react. Model predictive control (MPC) and artificial potential function (APF) are especially suitable for solving this problem. The advantage of MPC is it could deal with the non-linearity of vehicle dynamics and has the ability to solve optimization problems with multi-constraints and multi-objective functions efficiently \cite{29Carvalho}. APF is initially applied in robotics path planning. It establishes a repulsive field for the obstacle and an attractive field for the goal point. Then the path is generated by moving along the direction of gradient descent \cite{30Khatib}. The benefit of using APF is its fast computation which could be used in real-time. Based on these advantages, MPC and APF have been widely used in emergency collision avoidance, either separately or jointly.
\indentpar
MPC is mainly used for high-level multi-objective optimization and low-level trajectory tracking. Funke \textit{et al.} \cite{31Funke} and Hajiloo \textit{et al.} \cite{32Hajiloo} use model predictive control to address the conflict of control objectives among collision avoidance, vehicle stability and path tracking. Liu also uses it in \cite{33Liu} to approach a target position as quickly as possible while stabilizing vehicles in an unknown environment. For low-level trajectory tracking, reference trajectories are pre-designed by using polynomial curves \cite{34Shim, 10Sun, 35Howard}, Bezier Curves \cite{9Khattar}, sigmoid functions \cite{36Li} and geometry relationships \cite{7Choi, 6Cui}. MPC is required to track or approach the reference trajectory as accurately as possible without collision and violating vehicle dynamics constraints. One of the key points in using MPC is how to realize collision avoidance. Carvalho \textit{et al.} \cite{37Carvalho, 38Carvalho} use the signed distance between the ego vehicle and its surrounding obstacles and generates approximate linear constraints. In \cite{22Liu}, ego vehicles and surrounding obstacles are modeled as rectangles and the non-intersection criteria is transformed into constraints in MPC by using Farka’s lemma. Sathya \textit{et al.} \cite{39Sathya} set obstacles as a set of nonlinear inequalities which doesn’t require the computation of projections, thus, increasing the computational speed. Rather than setting collision avoidance as constraints, Gao \textit{et al.} \cite{40Gao} include collision avoidance cost in the cost function which is derived from the longitudinal distance from the ego vehicle and obstacles.
\indentpar
Generally, APF is used to generate reference paths for tracking methods \cite{43Lin, 44Liu} or set as a penalty term in the cost function for collision avoidance in optimization problems \cite{45Pongsathorn}. When combined with MPC, APF plays the same roles above and additional vehicle dynamics constraints will be included in the MPC settings, leading to better performance in practice. In \cite{21Ji, 41Rasekhipour, 42Huang}, APF will give the reference trajectory and a multi-constraint model predictive controller is used to track it. In \cite{22Liu, 19Wang, 23Cao}, APF is used as a penalty term in the cost function for MPC. The key point is how to design the APF. Exponential and trigonometric functions are implemented in \cite{46Wolf, 47Pozna} and the signed distance function is used in \cite{19Wang, 23Cao}. A novel harmonic velocity potential approach is used to address the local minima problem in \cite{48Cao}. Although a lot of work has been done in combining MPC and APF \cite{20Huang, 49Canale, 50Lin} including the work mentioned above, they have several common limitations and there is still room for improvement. The first limitation is the previous APFs for road structures only consider the straight or curved road and could not be extended in a more complicated environment such as an intersection, thus lack of generality. The second limitation is that previous APFs cannot realize collision mitigation which is important in an emergency as the collision might be inevitable. The third limitation is that the ego vehicle is usually considered as a mass point or a combination of circles which needs to set extra safe regions around obstacles and further decreases the possibility of finding a feasible solution. Motivated to address these three limitations, we propose a generalized safe controller by combining MPC and APF. We compare it with the current safe control method since both of them are designed for using in a generalized environment. 

\indentpar
Hamilton-Jacobi reachability (HJ-reachability) \cite{14Bansal} and control barrier function (CBF) \cite{15Ames} are the state-of-arts for synthesizing a safe controller. The core idea of these methods in collision avoidance is finding the control invariant set outside the avoid set (obstacle) so that the system will never enter into the avoid set if the control input of the system satisfies the constraints for the control invariant set. HJ-reachability usually uses the signed distance function between the state and the avoid set as its value function and the optimal control given from it will try to minimize the signed distance at the end of the time period. It could give the largest control invariant set, and deal with system uncertainty and input constraints. The CBF usually needs to be hand-crafted by experts or constructed from some machine learning methods \cite{16Srinivasan}. When taking the path planning problem as an optimization problem, both the CBF and the modified HJ-reachability would be set as hard constraints to avoid collisions. Some performance terms such as distance to the goal, and energy efficiency could be set as soft constraints to realize specific tasks \cite{15Ames}. However, for HJ-reachability and learning-based CBF, they will be computationally expensive for systems with high dimensions. In this case, the system needs to be decomposed to use HJ-reachability which will not guarantee safety for multiple avoid sets \cite{17Chen}. Also, these methods do not consider collision mitigation when the system is not in control invariant set \cite{18Leung}. 

\indentpar
Compared with previous work that combines APF and MPC, the innovations of our safe controller are 1. Utilizing a new potential function that considers the shape of the obstacle vehicle, ego vehicle, and road structure, enabling its use in various road structures; 2. Pre-assigning protection to the human-occupied part of the vehicle by adjusting the artificial potential function; 3. The method can be easily generalized for using in multi-vehicle cooperative control. Compared with the previous safe control method, the contributions of our safe controller are 1. Considering the combined effect of all obstacles rather than switching from tasks that individually avoid each obstacle, leading to improved performance; 2. Taking collision mitigation into consideration which could increase the survival rate in an inevitable collision. To the best of our knowledge, no previous work uses APF+MPC in the intersection or could realize specific part protection. For path planning in the intersection, the existing methods include choosing paths from pre-defined path libraries \cite{10Sun, 11Wei} and using the sample-based planning method\cite{12Wu, 13Xihui}. MPC is usually used to decide the order of vehicles for crossing the intersection \cite{56Georg}. For collision mitigation, the most similar work is \cite{19Wang} which constructs the damage severity as a cost function and the minimum damage is realized by adjusting the relative collision velocity and angle. 

\section{Algorithm Framework}
\subsection{Problem Statement}
\begin{figure*}[!ht]
\centering
\includegraphics[width=1\textwidth]{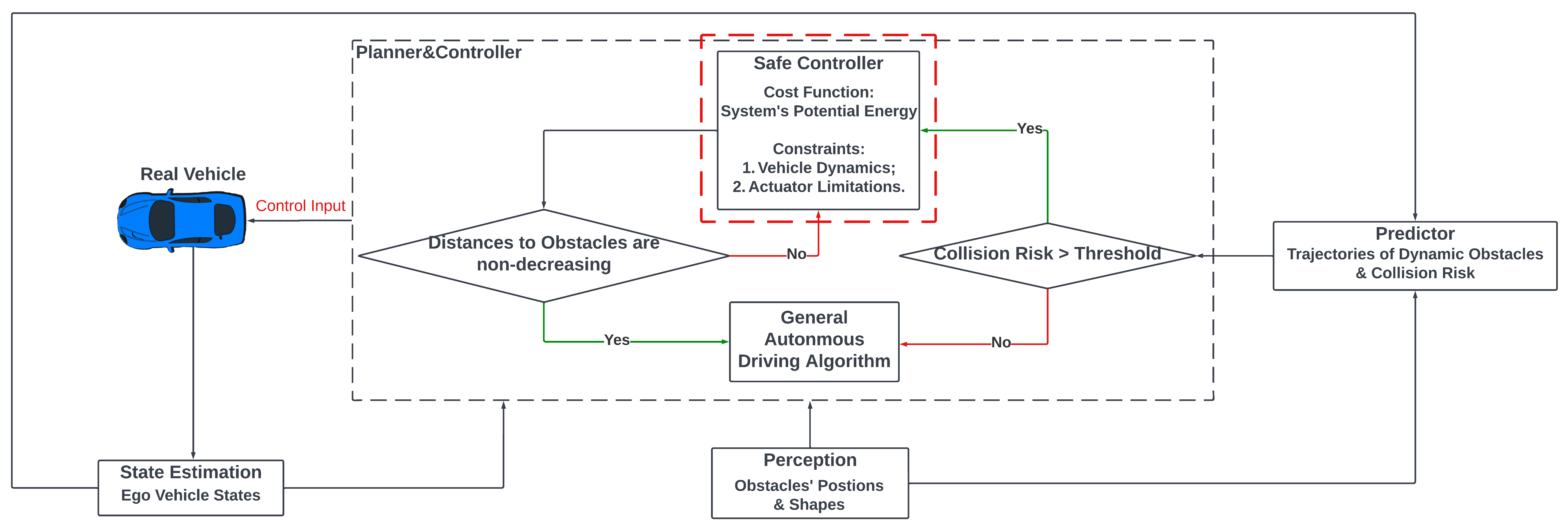}
\caption{Strategy Illustration}
\label{WF}
\end{figure*}

Although this paper focuses on the module of the safe controller, we list all the other modules for autonomous vehicles as shown in Fig. \ref{WF} to clarify our task. The planner will receive the ego vehicle's states, the static obstacles' positions, and trajectories of the dynamics obstacles from the estimation module, perception module, and predictor. Moreover, the predictor will send the collision risk for the planner to decide whether to enable the safe controller. Different indicators such as time to collision (TTC) \cite{51Jitendra, 52Seewald}, dynamic safety distance \cite{53He} and stochastic reachable set threat assessment \cite{57Khattar, 58Khattars} could be used in the predictor to decide the collision risk. 
\indentpar
Given all the information above, the safe controller seeks to find a safe path for the ego vehicle to get away from obstacle vehicles and realize the pre-assigned part(e.g., driver's position) protection. The safe controller will be activated until distances to obstacles are non-decreasing. The constraints are vehicle dynamics and actuator limitations obtained from vehicle models. The cost function is the potential energy of the system created by the ego vehicle and obstacles, which is calculated by placing the charged ego vehicle into the potential field generated by the obstacles.  
\subsection{Vehicle Models}
\begin{figure}[!ht]
\centering
\subfloat[]{\includegraphics[width=0.24\textwidth]{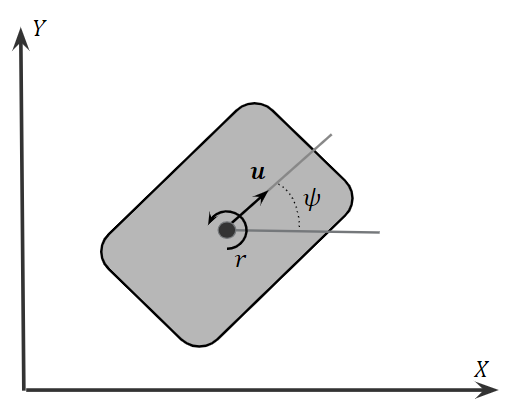}\label{Uni}}
\subfloat[]{\includegraphics[width=0.26\textwidth]{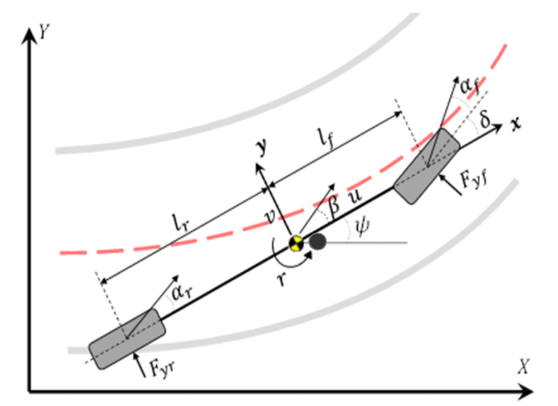}\label{DB}}\hskip1ex
\caption{Different Vehicle Models. (a) Unicycle Model. (b) Dynamic Bicycle Model. The figure is taken from \cite{42Huang}.}
\label{VM}
\end{figure}

We use two vehicle models in our work as shown in Fig. \ref{VM}. A unicycle model in Fig. \ref{Uni} is used for comparison with HJ-reachability because of the computational limitation. Then a dynamic bicycle model in Fig. \ref{DB} is used for validation.
\indentpar
The equations of motion of the unicycle model are:
\begin{equation}\label{UN1}
    \dot{X} = u \ cos(\psi)
\end{equation}
\begin{equation}
    \dot{Y} = u \ sin(\psi)
\end{equation}
\begin{equation}
    \dot{\psi} = r
\end{equation}
\begin{equation}\label{UN4}
    \dot{u} = a
\end{equation}
where $X$ is the longitudinal position, $Y$ is the lateral position, $\psi$ is the orientation and $u$ is the speed of the vehicle. The angular velocity and acceleration of the vehicle are represented as $r$ and $a$. 
\indentpar
The equations of motion of the dynamic bicycle model are \cite{42Huang, 19Wang, 41Rasekhipour}:
\begin{equation}\label{DB1}
    \dot{X} = u \ cos(\psi) - v \ sin(\psi)
\end{equation}
\begin{equation}
    \dot{Y} = v \ cos(\psi) + u \ sin(\psi)
\end{equation}
\begin{equation}
    \dot{\psi} = r
\end{equation}
\begin{equation}
    m(\dot{u}-vr) = F_{xT}
\end{equation}
\begin{equation}
    m(\dot{v} + ur) = F_{yf} + F_{yr}
\end{equation}
\begin{equation}\label{DB6}
    I_{z}\dot{r} = l_f \ F_{yf} - l_r \ F_{yr}
\end{equation}
where $X$, $Y$, and $\psi$ are the longitudinal position, lateral poison and orientation of the vehicle in the global frame, $u$, $v$ and $r$ represent the longitudinal velocity, lateral velocity, and angular velocity of the vehicle at its local frame whose origin is at the center of gravity, $m$ and $I_z$ denote the mass and the inertia of the vehicle, $F_{yf}$ and $F_{yr}$ are the lateral forces of the front and rear tires and $F_{xT}$ is the total longitudinal force of tires. Using the linear tire model, the lateral tire forces are:
\begin{equation}
    F_{yf} = C_f \alpha_f = C_f (\delta - \frac{v+l_f r}{u})
\end{equation}
\begin{equation}\label{DB8}
    F_{yr} = C_r \alpha_r = C_r (-\frac{v-l_r r}{u})
\end{equation}
where $\delta$ is the steering angle and $C_f$ and $C_r$ are cornering stiffness values of the front and rear tires, $\alpha_f$ and $\alpha_r$ are the sideslip angles of the front and rear tires.

\subsection{Potential Function}
\begin{figure}[!ht]
\vspace*{-0.4cm}
\centering
\includegraphics[width=0.3\textwidth]{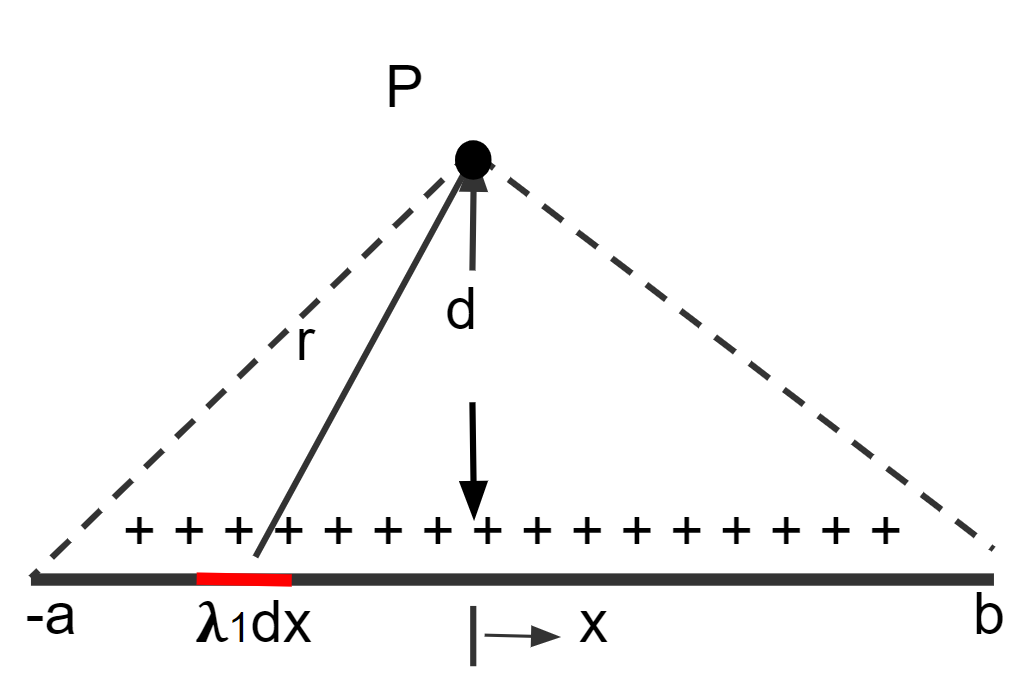}
\caption{Finite-long line charge}
\label{FLLC}
\end{figure}

Assuming that the goal of the optimal control is to minimize the cost function, then APF is designed to become larger at the place closer to obstacles. Thus, the ego vehicle would tend to get away from obstacles when APF is set as a penalty term which realizes collision avoidance. We design a new potential function for our problem, and it is inspired by the electrical potential field of line charge. Two basic elements of the potential function are finite-long line charge and infinite-long line charge. 
\indentpar
The potential field of the point $P$ around the finite-long line charge could be derived as ~\cite{2EF}: 
\begin{equation}\label{FLEQ}
\begin{multlined}
V_{F} = \int_{-a}^{b} \frac{k \lambda_1 dx}{r} = \int_{-a}^{b} \frac{k\lambda_1 dx}{\sqrt{x^2+d^2}} \\
=  k\lambda_1 ln(\frac{b+\sqrt{b^2 + d^2}}{-a + \sqrt{a^2+d^2}}).
\end{multlined}
\end{equation}
where $k$ is the coulomb constant, $\lambda_1$ is the line density of the charger, $a$ is the lower bound of the line, $b$ is the upper bound of the line and $d$ is the distance between point $P$ and the charged finite line as shown in Fig. \ref{FLLC}.

\begin{figure}[!ht]
\centering
\includegraphics[width=0.4\textwidth]{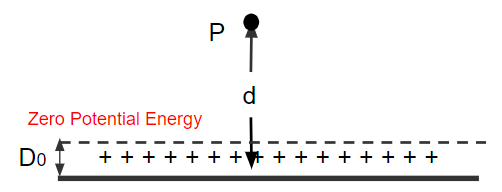}
\caption{Infinite-long line charge}
\label{ILLC}
\end{figure}

\begin{figure}[!ht]
\centering
\includegraphics[width=0.47\textwidth]{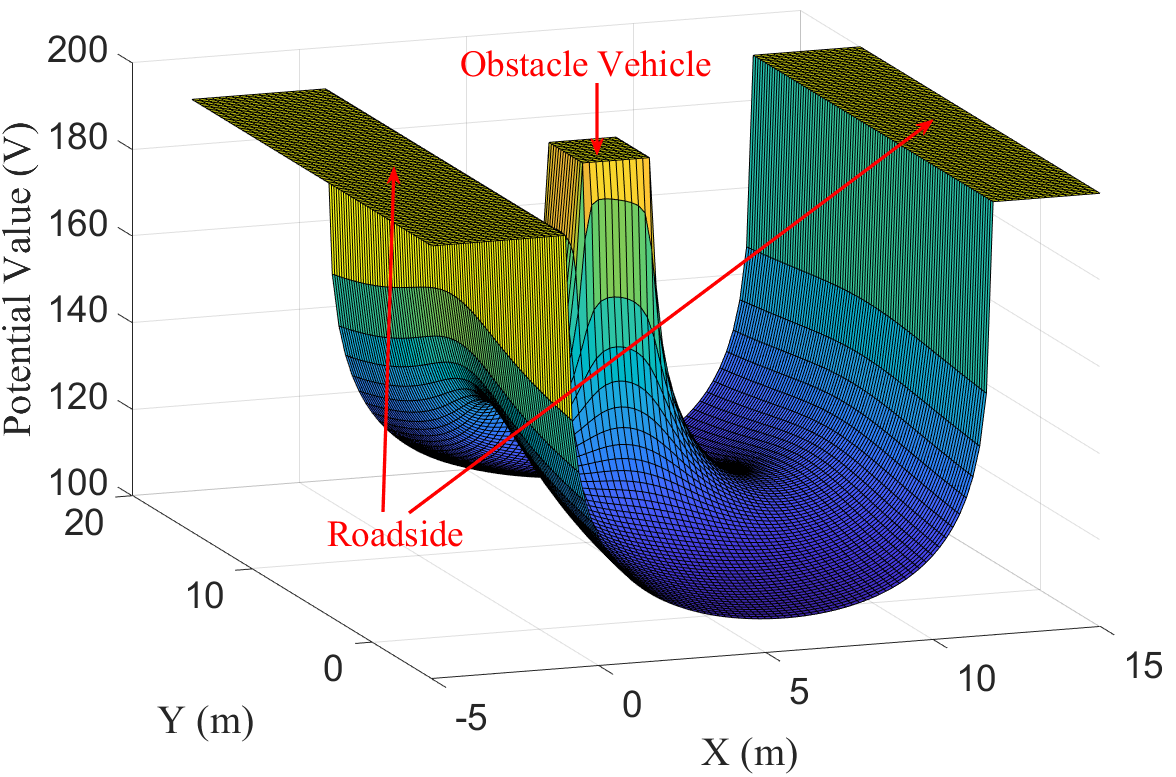}
\caption{Potential Field in highway}
\label{PFH}
\end{figure}

\begin{figure}[!ht]

\centering
\includegraphics[width=0.47\textwidth]{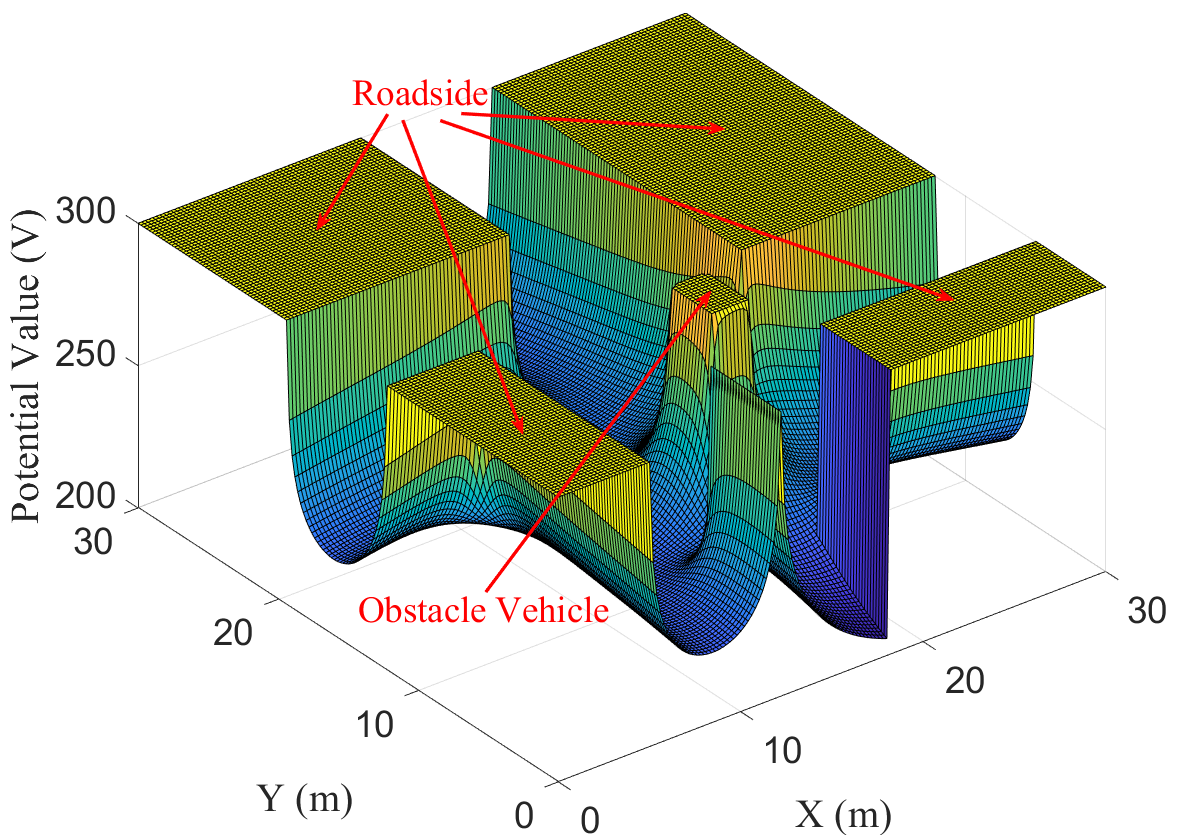}
\caption{Potential field in intersection}
\label{PFI}
\end{figure}

The potential field of the point $P$ around an infinite-long line charge could be derived as ~\cite{2EF}:
\begin{equation}\label{IFLEQ}
V_{Inf} = 2k\lambda_1 ln(\frac{d_0}{d}),
\end{equation}
where $k$ is the coulomb constant, $\lambda_1$ is the line density of the charger, $d$ is the distance between point $P$ and the infinite-long line charge and $d_0$ is the zero potential energy distance which is shown in Fig. \ref{ILLC}.
\indentpar
For the obstacle vehicle, since its shape is a rectangle in 2D, its potential field is constructed by the superposition of potential fields of four finite-long line charges. The road structure in the highway or the normal road is considered as the infinite-long line charge while it is considered as the semi-infinite-long charge in the intersection. The overall potential field is the superposition of the potential field of all surrounding obstacles. It could be written as $V = \sum_{i = 1:n} w_i V_{Obs_i}$ where $w_i$ is the weight and $V_{Obs_i}$ is the potential function of $i$th obstacle. The potential fields for highway and intersection with one obstacle vehicle are shown in Fig. \ref{PFH} and Fig. \ref{PFI}. 

\subsection{Potential Energy}
\begin{figure}[!ht]
\centering
\includegraphics[width=0.4\textwidth]{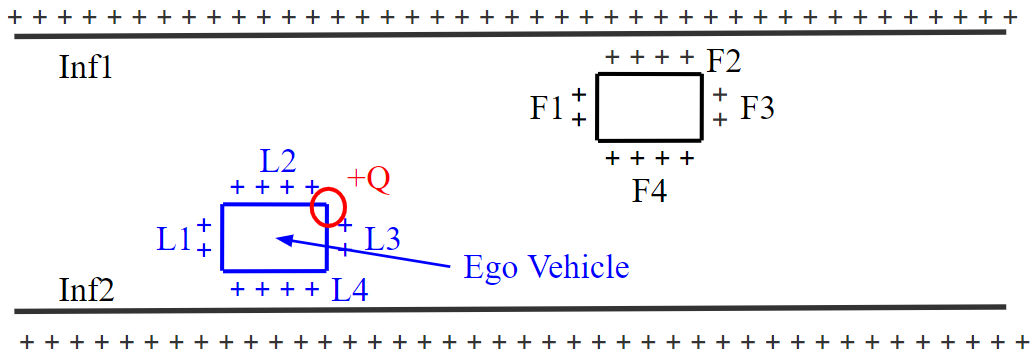}
\caption{Charged System}
\label{PE}
\end{figure}

\begin{figure}[!ht]
\centering
\includegraphics[width=0.4\textwidth]{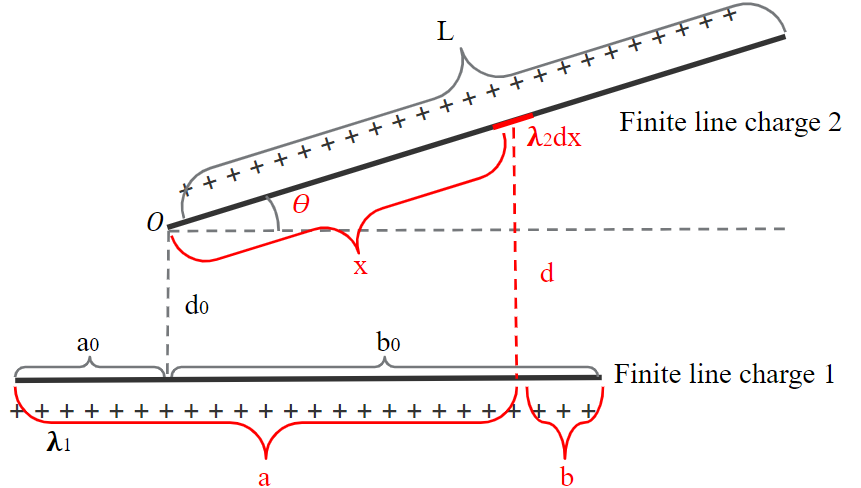}
\caption{Finite line charges}
\label{SysLine}
\end{figure}

Our method calculates the system's potential energy by placing the charged ego vehicle into the potential field constructed by surrounding obstacles. Fig. \ref{PE} is an example to show our calculation. The ego vehicle is on the left and is constructed by four finite-long line charges. An extra point charge is placed at the driver's position to realize the specific part protection.
The potential energy of the system could be calculated as:
\begin{equation}
    E = E_R + E_V + E_{pR} + E_{pV}
\end{equation}
where $E_R$ and $E_V$ represent the potential energy between the ego vehicle and road structures and the obstacle vehicles, and $E_{pR}$ and $E_{pV}$ represent the potential energy between the extra point charge and road structures and the obstacle vehicles. After rearranging, it could be written as:
\begin{equation}\label{PEEQ}
\begin{split}
    E = &\sum_{i=1:4}\overbrace{(\sum_{j=1:l}E_{L_i F_j} + \sum_{m=1:p}E_{L_i Inf_k})}^{\text{Term1}} + \\
    &\underbrace{(\sum_{j=1:l}E_{p F_j} + \sum_{m=1:p}E_{p Inf_k})}_{\text{Term2}}
\end{split}
\end{equation}
where $\text{Term1}$ represents the potential energy between each finite line charge of the ego vehicle and the finite and infinite line charge of the surrounding obstacles, and $\text{Term2}$ represents the potential energy between the extra point charge and the finite and infinite line charge of the surround obstacles. Parameters $l$ and $m$ represent the number of finite and infinite line charges contained in the surrounding obstacles. In this example, $l = 4$ and $m = 2$. 
\indentpar
The detailed calculation of these four kinds of potential energy is presented below:
\indentpar
1) \textit{Two finite line charges}: The potential energy between two finite line charges could be calculated as:
\begin{equation}\label{FFEQ}
\begin{split}
    E_{L_i F_j} = &\int_0^L V_F \lambda_2 dx \\
    = &\int_0^L k\lambda_1 \lambda_2 ln(\frac{b+\sqrt{b^2 + d^2}}{-a + \sqrt{a^2+d^2}})dx
\end{split}
\end{equation}
As shown in Fig. \ref{SysLine}, $\lambda_2$ and $L$ are the line densities and length of line charge 2. The definition of $\lambda_1$, $a_0$, $b_0$ and $d_0$ are the same as Fig. \ref{FLLC} with respect to point $O$. Then, we could represent $a$, $b$ and $d$ as: 
\[
a = a_0 + x \ cos(\theta), \ b = b_0 - x \ cos(\theta), \ d = d_0 + x \ sin(\theta).
\]
Thus, $E_{L_i F_j}$ is an integration about variable $x$ with known parameters $k$, $\lambda_1$, $\lambda_2$, $a_0$, $b_0$, $d_0$, $\theta$ and $L$. To the best of our knowledge, there is no existing work given the explicit solution for this integration. MATLAB and WolframAlpha also fail to solve it explicitly. The numerical method is used during the implementation. 

\begin{figure}[!ht]
\centering
\includegraphics[width=0.4\textwidth]{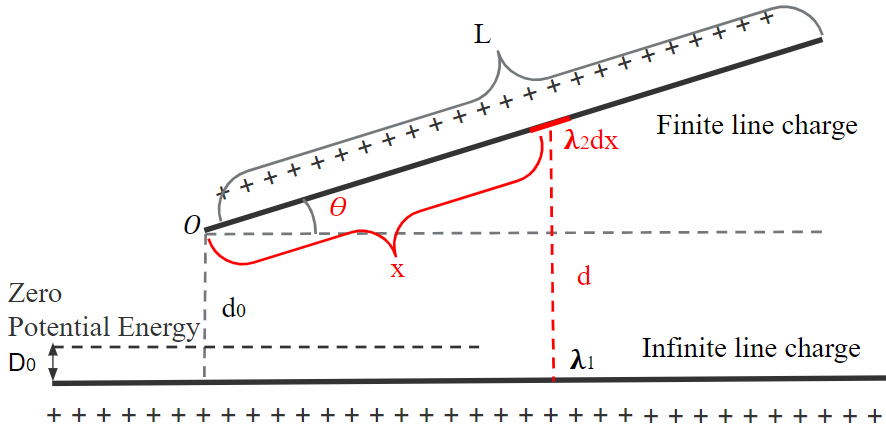}
\caption{Finite line charges}
\label{FINFLine}
\end{figure}

2) \textit{Finite line charge and infinite line charge}: The potential energy between finite line charge and infinite line charge could be calculated as:
\begin{equation}\label{FIEQ}
\begin{split}
    E_{L_i Inf_k} = &\int_0^L V_{Inf} \lambda_2 dx \\
    = &\int_0^L 2k\lambda_1 \lambda_2 ln(\frac{D_0}{d})dx
\end{split}
\end{equation}
 where $d = d_0 + x \ sin(\theta)$ and the definition of other parameters in Fig. \ref{FINFLine} is the same as Fig. \ref{SysLine} and Fig. \ref{ILLC}. It is also an integration for variable $x$ and the explicit result is: 
 \begin{equation}
 \begin{split}
     &E_{L_i Inf_k} = C \ (\left. x + ln(\frac{D_0}{d_0 + x \ sin(\theta)})(x+\frac{d_0}{sin(\theta)}) \right \vert_{0}^L) \\
     &= C \ (L + ln(\frac{D_0}{d_0+L\ sin(\theta)})(L+ \frac{d_0}{sin(\theta)})\\
     & -ln(\frac{D_0}{d_0})(\frac{d_0}{sin(\theta)}))
\end{split}
 \end{equation}
 where $C = 2k\lambda_1 \lambda_2$.

3) and 4) \textit{Point charge and line charge(finite $\&$ infinite)}: The potential energy of a point charge could be calculated as $E = Vq$ in which $V$ represents the potential, and $q$ represents the quantity of the charge. Thus, using (\ref{FLEQ}) and (\ref{IFLEQ}), the potential energy between the point charge and finite line charge is: 
\begin{equation}\label{QFEQ}
    E_{qF_j} = Q \ V_F = Q \ k\lambda_1 ln(\frac{b+\sqrt{b^2 + d^2}}{-a + \sqrt{a^2+d^2}})
\end{equation}
and the potential energy between the point charge and infinite line charge is:
\begin{equation}\label{QIEQ}
    E_{qInf_k} = Q \ V_{Inf} = 2Qk\lambda_1 ln(\frac{d_0}{d})
\end{equation}
\indentpar
The potential energy of the whole system in (\ref{PEEQ}) could be calculated by combing (\ref{FFEQ})-(\ref{QIEQ}). Noting that: 1. The potential energy of the whole system should also include the potential energy between obstacles. We don't discuss and include it here since it will not be affected by the decision variables in the optimization problem; 2. When two line charges intersect, (\ref{FFEQ}) and (\ref{FIEQ}) are no longer valid and the potential energy becomes infinite.   
\subsection{MPC Formulation}
The emergency collision avoidance problem could be formulated as a nonlinear model predictive control problem:
\begin{equation}
\underset{z_k,u_k}{min} \ \sum\limits_{k=1}^N E(z_k)
\end{equation}
\[
s.t. \ \ z_{k+1}= z_k + f(z_k, u_k)T
\]
\[
u_{min} \le u_k \le u_{max}.
\]
\[
k = 0,...,N-1
\]
where $E(z_k)$ represents the potential energy of the system which is (\ref{PEEQ}), $N$ is the prediction horizon and $T$ is the time step.
\indentpar
For the unicycle model, $z_k = [X; Y; \psi; u]$ and $f(z_k, u_k)$ could be derived by combing (\ref{UN1})-(\ref{UN4}). Inputs are $[r; a]$ and $u_{min}$ and $u_{max}$ represent the lower and upper limits of the angular velocity and the acceleration.
\indentpar
For the dynamic bicycle model, $z_k = [X;Y; \psi; u; v; r]$ and $f(z_k, u_k)$ could be derived by combing (\ref{DB1})-(\ref{DB8}). Inputs are $[F_{xT}; \delta]$ and $u_{min}$ and $u_{max}$ represent the lower and upper limits of the total longitudinal tire force and the steering angle.
\indentpar
We try to minimize the system's potential energy without violating vehicle dynamics and actuator limits listed above. Different obstacles and the ego vehicle could be approximated as polygons and then represented by the combination of finite line charge and infinite line charge, as in Fig. \ref{PE}. Values of $l$ and $p$ will also change accordingly. Although a simpler rectangular shape is used in Fig. \ref{PE} for illustration, the method is generic enough to be used for more complex shapes with increasing computational complexity. Thus, our method could consider the shapes of both obstacles and the ego vehicle. Moreover, the specific part protection could be realized by adding the additional point charge to the pre-assigned part. The penalty for the specific part of getting close to obstacles will be larger than other parts due to the existence of the point charge. That makes the collision less likely to happen at the specific part.  

\subsection{HJ-reachability}

A brief introduction about HJ-reachability is provided in (\ref{HJIntro}). For a detailed explanation of HJ-reachability, please refer to \cite{14Bansal}.
In this part, we will mainly introduce system decomposition in HJ-reachability. 
\indentpar
The total dimension of our system is 8 for one obstacle vehicle and 12 for two obstacle vehicles which are not computationally possible even offline \cite{14Bansal}. Thus, we decompose it into two to three subsystems based on the obstacle vehicle's number. We use one obstacle vehicle's case to illustrate the decomposition. We decompose the whole system whose dimension is 8 into subsystem one and subsystem two. Subsystem one is constructed by taking the ego vehicle as the reference and calculating the relative position of the obstacle vehicle and the ego vehicle. This subsystem is used to avoid collision between the ego and obstacle vehicles. The dimension of subsystem one is 6 and the state-space representation of this system is:
\begin{equation}
\dot{z}= 
\begin{bmatrix}
\dot{x_{rel}} \\
\dot{y_{rel}}\\
\dot{\psi_{Ego}} \\
\dot{\psi_{Obs}} \\
\dot{u_{Ego}} \\
\dot{u_{Obs}} 
\end{bmatrix}
+
\begin{bmatrix}
u_{Obs} cos(\psi_{Obs}) - u_{Ego} cos(\psi_{Ego}) \\
u_{Obs} sin(\psi_{Obs}) - u_{Ego} sin(\psi_{Ego})\\
r_{Ego} \\
r_{Obs} \\
a_{Ego} \\
a_{Obs}
\end{bmatrix}
\end{equation}
where $x_{rel}$ and $y_{rel}$ denote the relative position of the ego and obstacle vehicles in the global frame, $\psi$, $u$, $r$ and $a$ are defined the same as the unicycle model for the obstacle vehicle and the ego vehicle. The grid size is $21\times31\times11\times11\times11\times11$ uniformly over the 6D system space $[x_{rel}; y_{rel}; \psi_{Ego}; \psi_{Obs}; u_{Ego}; u_{Obs}] \in [-15,15]\times[-20,20]\times[-\pi,\pi]\times[-\pi,\pi]\times[0,25]\times[0,25]$. Subsystem two is used for avoiding collision between the ego vehicle and the road structure and its state-space representation is: 
\begin{equation}\label{DCUNI}
\dot{z}
=
\begin{bmatrix}
\dot{x_{Ego}} \\
\dot{y_{Ego}}\\
\dot{\psi_{Ego}} \\
\dot{u_{Ego}}
\end{bmatrix}
= 
\begin{bmatrix}
u_{Ego} cos(\psi_{Ego}) \\
u_{Ego} sin(\psi_{Ego})\\
r_{Ego} \\
a_{Ego}
\end{bmatrix}.
\end{equation}
The notations in (\ref{DCUNI}) are defined the same as the unicycle model for the ego vehicle. The grid size is $41\times41\times21\times41$ uniformly over the 4D system space $[x_{Ego}; y_{Ego}; \psi_{Ego}; u_{Ego}] \in [-5,25]\times[-5,25]\times[-\pi,\pi]\times[0,40]$. We use the signed distance function as the value function $V$ for both subsystem one and subsystem two. The value function for the whole system is calculated as $V = min\{V_1, V_2\}$ where $V_1$ and $V_2$ are value functions of two subsystems and the selection of control input depends on the corresponding subsystem.

\section{Simulation Results}
We test 192 cases from 8 different scenarios and compare the results of our method with results obtained from HJ-reachability. Scenarios are described in the following table(Table \ref{tabS}).

\begin{table*}
\vspace{-0.5em}
\begin{center}
\caption{Scenario Description}
\label{tabS}
\begin{tabular}{|m{0.25\textwidth}|m{0.2\textwidth}|m{0.25\textwidth}|m{0.2\textwidth}|}
\hline
\textbf{Scenario} & \textbf{Description} & \textbf{Scenario} & \textbf{Description} \\
\hline
Scenario 1:
\centering
\includegraphics[width=0.25\textwidth]{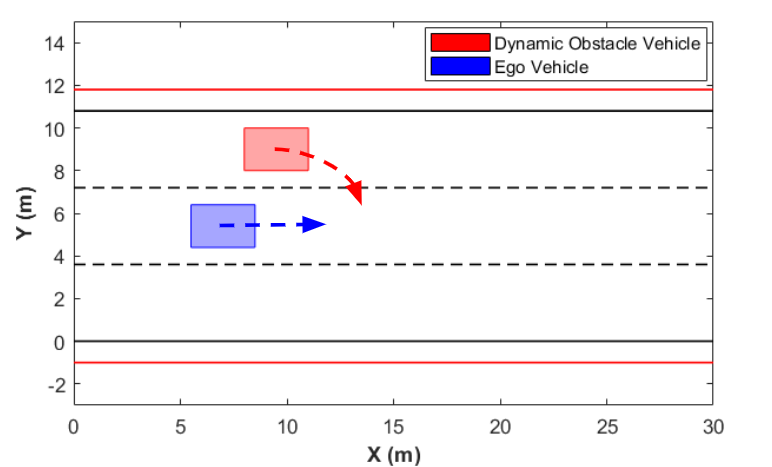}
&It is a three lanes highway scenario with one obstacle vehicle. The obstacle vehicle makes a dangerous lane change and loses control. It will keep turning right to the roadside until it gets out of the road.
&Scenario 2:
\centering
\includegraphics[width=0.25\textwidth]{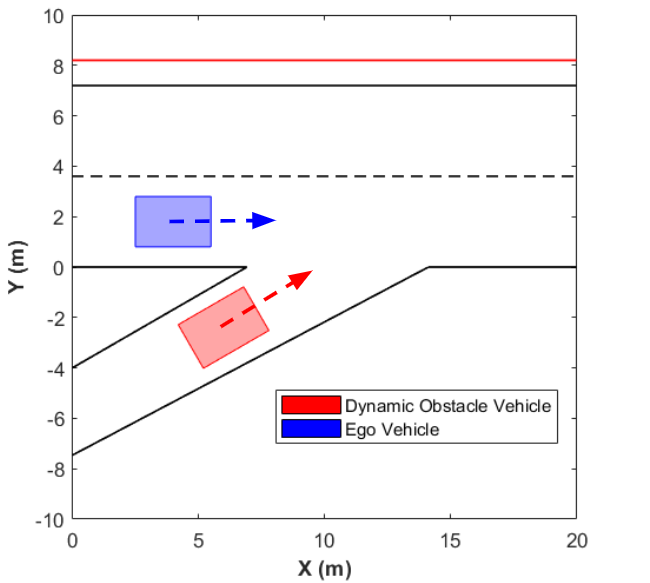}
&It is a highway merging scenario with one obstacle vehicle. The obstacle vehicle tries to merge to the highway from the right lane but drives too fast or loses control so that it moves across lanes rather than merging into the highway.   \\
\hline
Scenario 3:
\centering
\includegraphics[width=0.25\textwidth]{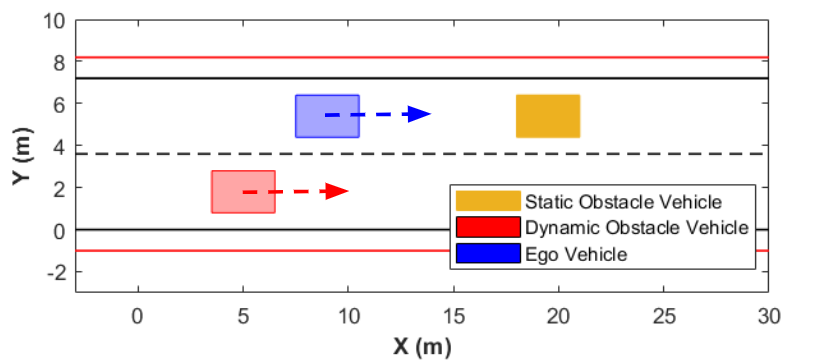}
& It is a two lanes highway scenario with two obstacle vehicles. Obstacle vehicle one will stay still in front of the ego vehicle in the left lane and obstacle vehicle two will move at a constant speed in the right lane. The ego vehicle needs to change to the right lane to avoid collision with obstacle vehicle one while avoiding collision with obstacle vehicle two.   
&Scenario 4:
\centering
\includegraphics[width=0.25\textwidth]{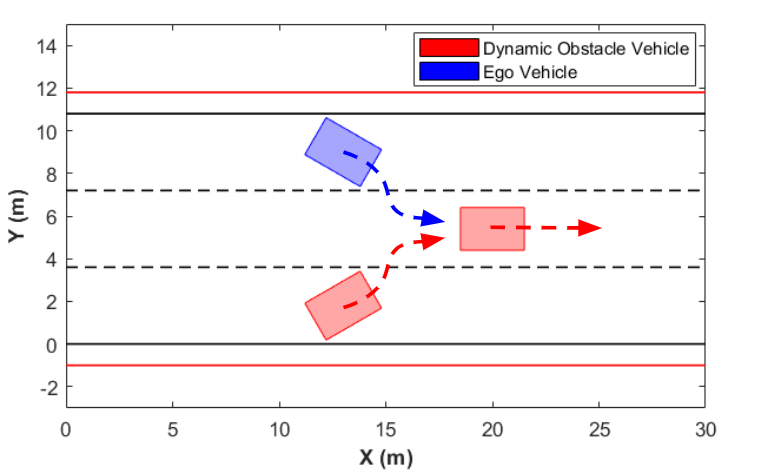}
&It is a three lanes highway scenario with two obstacle vehicles. Obstacle vehicle one will move at a constant speed in the middle lane and obstacle vehicle two will try to move to the middle lane from the right lane. Both obstacle vehicle two and ego vehicle plan to turn to the middle lane and occupy the same position behind obstacle vehicle one.\\
\hline
Scenario 5:
\centering
\includegraphics[width=0.25\textwidth]{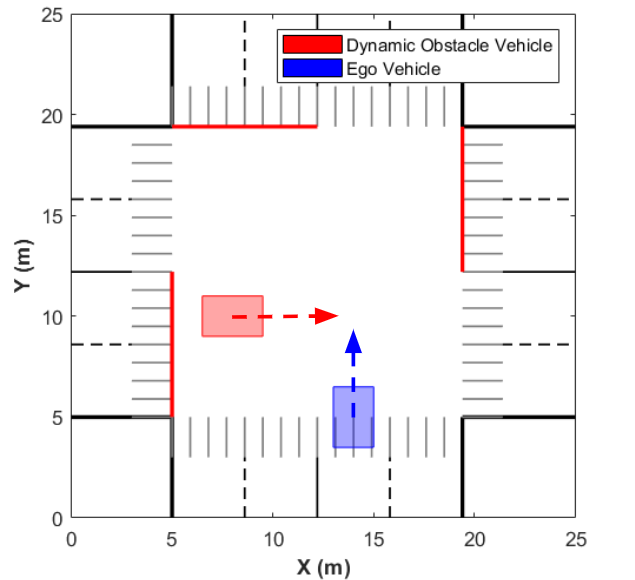}
&It is an intersection scenario with one obstacle vehicle. The ego vehicle intends to move straight to pass this intersection from south to north. The obstacle vehicle breaks the traffic rule and tries to move across the intersection at the red light from west to east. If no action is taken, it will be a severe “T-bone” collision.
&Scenario 6:
\centering
\includegraphics[width=0.25\textwidth]{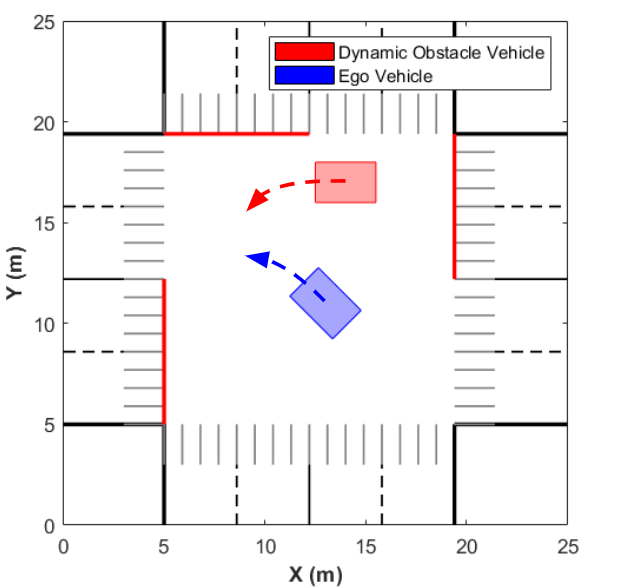}
&It is an intersection scenario with one obstacle vehicle. The ego vehicle intends to turn left from the south to the west. The obstacle vehicle breaks the traffic rule and tries to turn left at the red light from east to south. The ego vehicle needs to make the decision to keep turning left or make a U-turn.\\
\hline 
Scenario 7:
\centering
\includegraphics[width=0.25\textwidth]{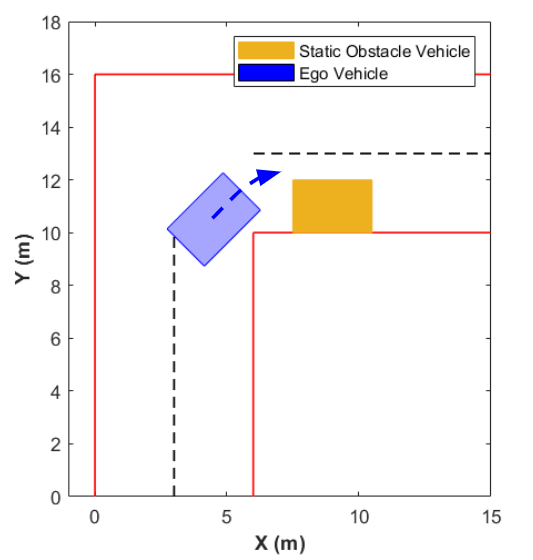}
&It is a turning scenario in normal road conditions with one obstacle vehicle. The ego vehicle suddenly finds a still obstacle vehicle that stays in its lane during the turning because of the blind spot. The ego vehicle needs to realize collision avoidance in a limited space. 
&Scenario 8:
\centering
\includegraphics[width=0.25\textwidth]{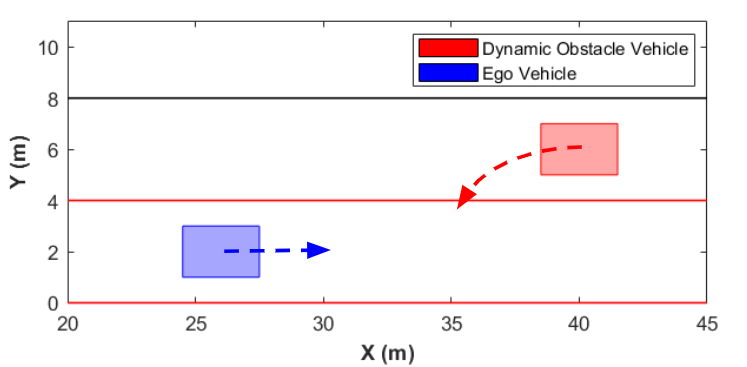} 
&It is a two lanes heading towards scenario with one obstacle vehicle. The ego vehicle and the obstacle vehicle drive towards each other but the obstacle vehicle loses control which makes it turn left to the ego vehicle’s lane.\\
\hline 
\end{tabular}
\end{center}
\end{table*}
\subsection{Results}
\begin{figure}[!ht]
\centering
\includegraphics[width=0.5\textwidth]{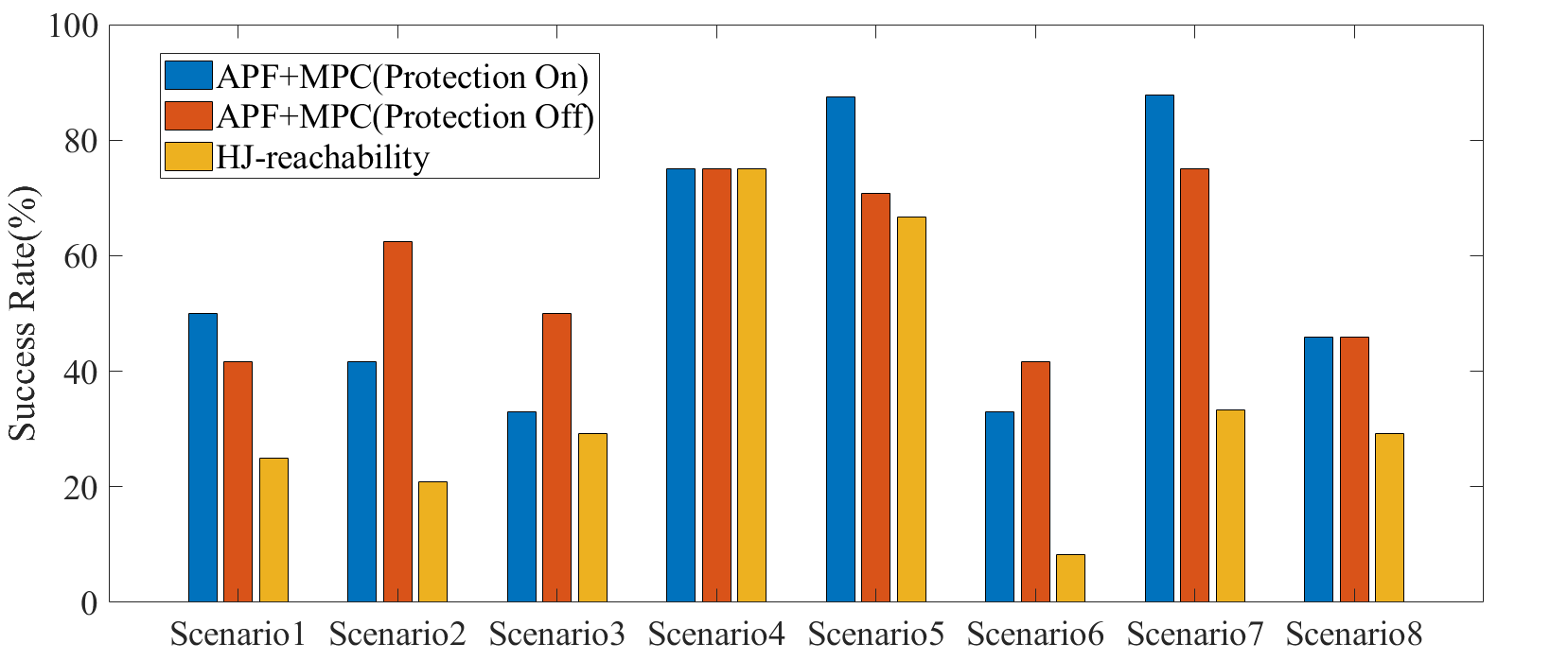}
\caption{Success rate comparison(Unicycle Model)}
\label{SRC}
\end{figure}

\begin{figure}[!ht]
\vspace{-0.2cm}
\centering
\includegraphics[width=0.5\textwidth]{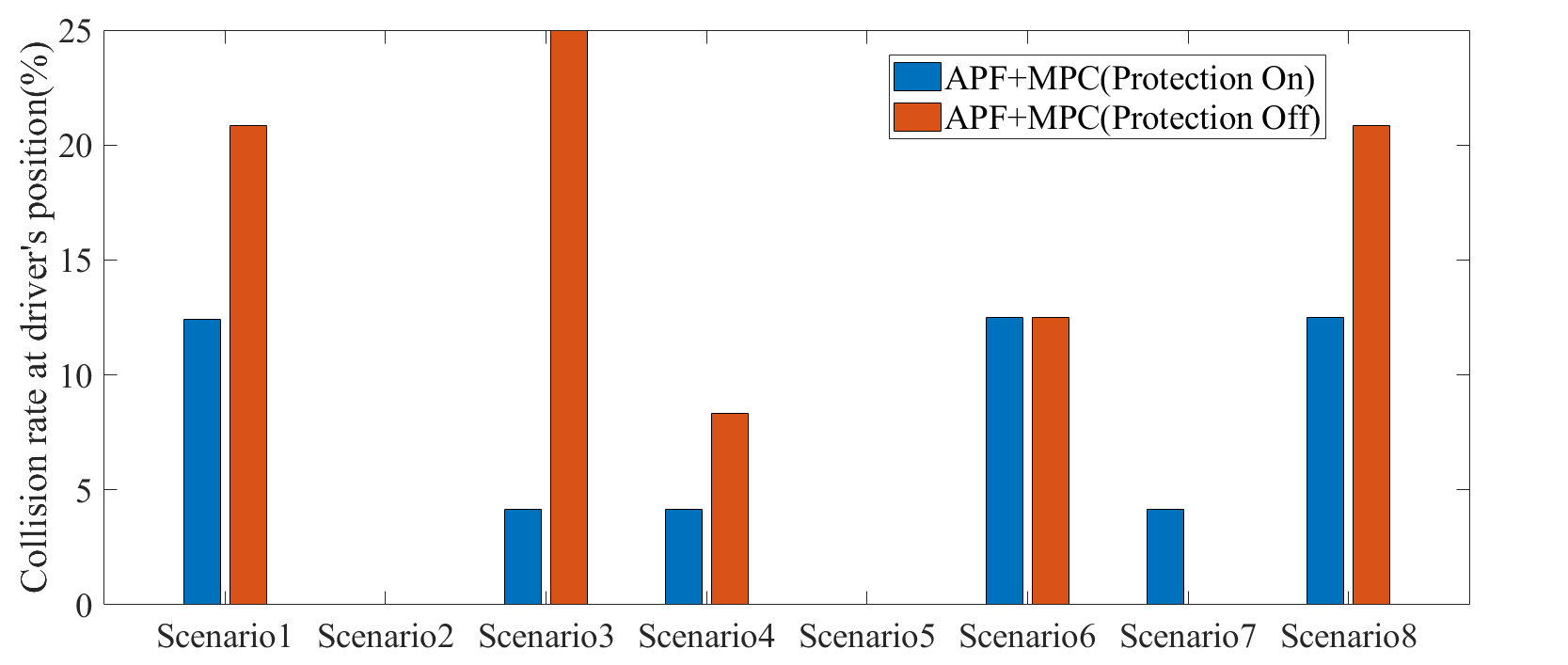}
\caption{Collision rate at driver's position comparison(Unicycle Model)}
\label{CRDP}
\end{figure}

\begin{figure}[!ht]
\centering
\includegraphics[width=0.5\textwidth]{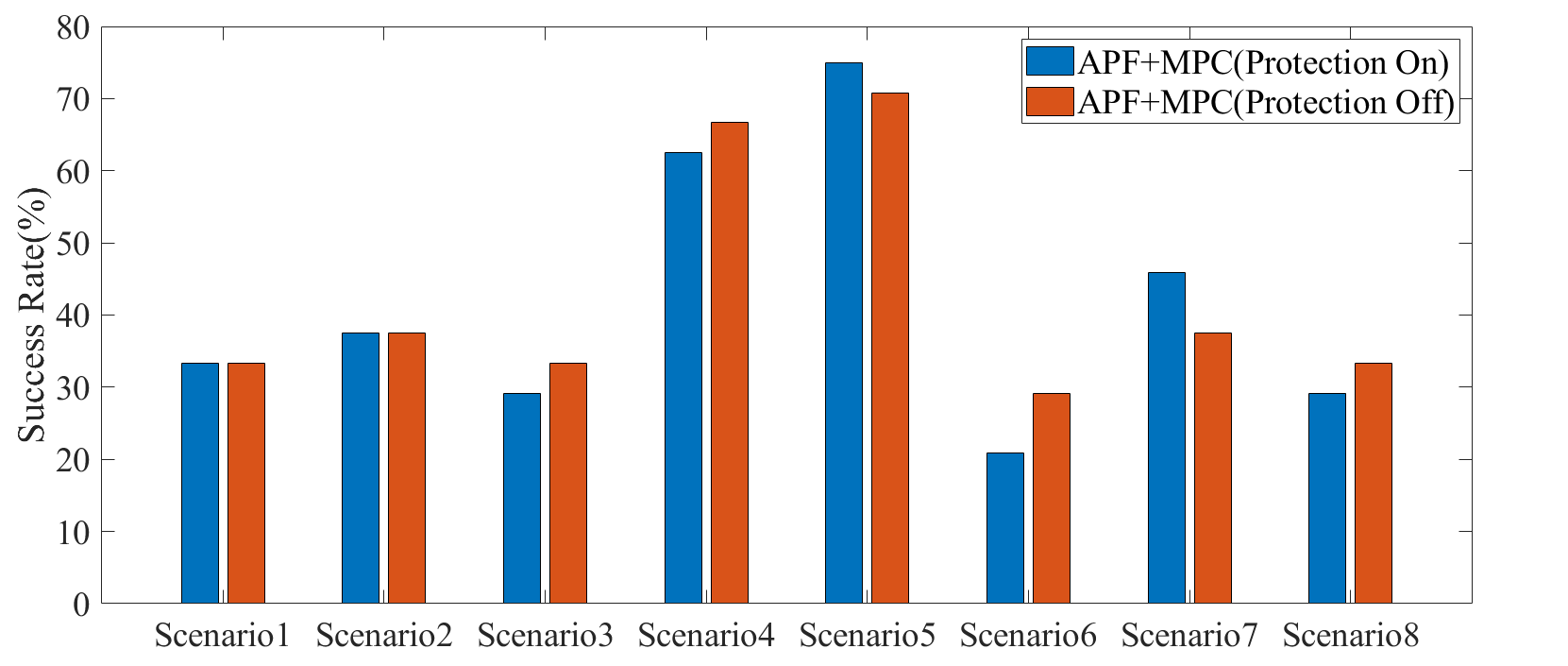}
\caption{Success rate comparison(Dynamic Bicycle Model)}
\label{SRCDB}
\end{figure}

\begin{figure}[!ht]
\centering
\includegraphics[width=0.5\textwidth]{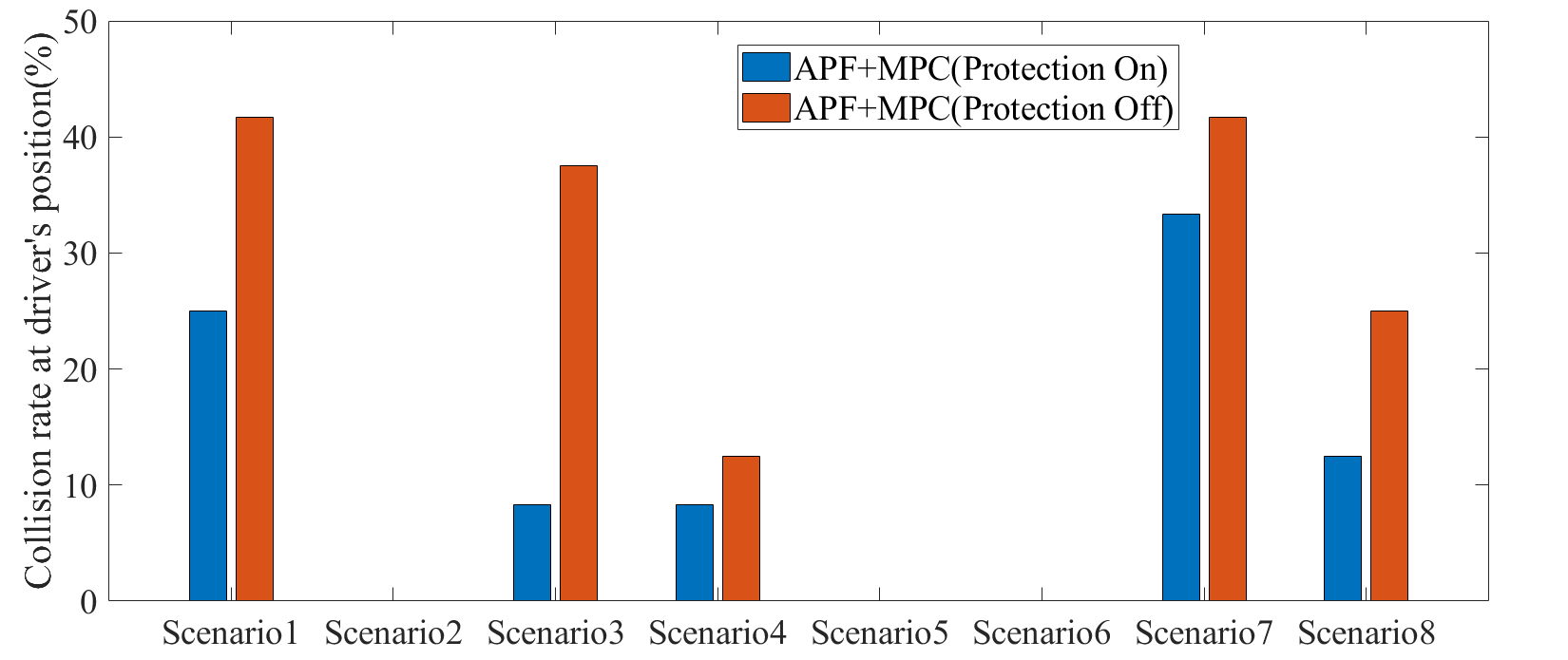}
\caption{Collision rate at driver's position comparison(Dynamic Bicycle Model)}
\label{CRDPDB}
\end{figure}

The results and initial states used in the test are shown in Table \ref{tab1}, Table \ref{tab2} and Fig. \ref{SRC}-\ref{CRDPDB}. In each test, the ego vehicle will start moving from its initial state until distances between the ego vehicle and its surrounding obstacles keep non-decreasing. If no collision happens during this period, we conclude that the ego vehicle successfully avoided the collision.
\indentpar
In using the unicycle model for comparison, the average success rate for APF+MPC when protection mode is on is 56.71$\%$ and is 57.81$\%$ when protection mode is off. Both of them are around 20$\%$ higher than the success rate of the HJ-reachability which is 35.94$\%$. From Fig. \ref{SRC}, we could see the success rate by using APF+MPC is also higher than HJ-reachability in each individual scenario except for scenario 4 in which their success rates are the same. Table \ref{tab2} and Fig. \ref{CRDP} show the effect of enabling driver protection mode. The average collision rate at the driver's position is 10.94$\%$ with protection mode off and 6.24$\%$ with protection mode on. Combining looking at Table \ref{tab1} and Table \ref{tab2}, we could see enabling the protection mode decrease 43$\%$ of collision that happens at the driver's position by decreasing 2$\%$ of the success rate which is a little sacrifice.
\begin{table*}
\begin{center}
\caption{Success rate comparison(Unicycle Model)}
\label{tab1}
\begin{tabular}{| c | c | c | c | c | c | c |}
\hline
Scenario & APF + MPC & APF + MPC & HJ Reachability & Velocity Range(mph) & Initial Relative Position & Initial Relative Position \\
 & (Protection On) & (Protection Off)& & & X-Direction(m)& Y-Direction(m)\\
\hline
1 & 50$\%$ & 41.67$\%$ &  25$\%$ &[45, 80] &3.6 & [-2, 4]\\
\hline 
2 & 41.67$\%$ & 62.5$\%$ &  20.83$\%$ &[45, 80] &4.8 & [-4, 2]\\
\hline
3 & 33$\%$ & 50$\%$ &  29.17$\%$ & [45, 80] & 3.6 & [-5, 3]\\
\hline 
4 & 75$\%$ & 75$\%$ &  75$\%$ &[45, 80] & 7.2 & [-1, 3]\\
\hline
5 & 87.5$\%$ & 70.83$\%$ &  66.67$\%$ &[20, 45] &[-6, -5] & [-5, -3]\\
\hline
6 & 33$\%$ & 41.67$\%$ &  8.33$\%$ &[20, 45] &[-3, -1] & [-5, -2]\\
\hline 
7 & 87.75$\%$ & 75$\%$ &  33.33$\%$ &[20, 55] & [-8, -5]& [-1, 1]\\
\hline
8 & 45.83$\%$ & 45.83$\%$ &  29.17$\%$ &[30, 70] &4 & [-10, -18]\\
\hline
\end{tabular}
\end{center}
\end{table*}

\begin{table}
\begin{center}
\caption{Collision rate at driver's position comparison(Unicycle Model)}
\label{tab2}
\begin{tabular}{| c | c | c |}
\hline
Scenario & APF + MPC & APF + MPC  \\
 & (Protection On) & (Protection Off)\\
\hline
1 & 12.5$\%$ & 20.83$\%$ \\
\hline 
2 & 0$\%$ & 0$\%$ \\
\hline
3 & 4.17$\%$ & 25$\%$ \\
\hline 
4 & 4.17$\%$ & 8.33$\%$ \\
\hline
5 & 0$\%$ & 0$\%$ \\
\hline
6 & 12.5$\%$ & 12.5$\%$ \\
\hline 
7 & 4.17$\%$ & 0$\%$ \\
\hline
8 & 12.5$\%$ & 20.83$\%$ \\
\hline
\end{tabular}
\end{center}
\end{table}

\indentpar
In using the dynamic bicycle model for validation, the average success rate for APF+MPC when protection mode is on is 41.67$\%$ and is 42.71$\%$ when protection mode is off. The average collision rate at the driver's position is 19.79$\%$ with protection mode off and 10.94$\%$ with protection mode on. Enabling the protection mode decreases 44.72$\%$ of collision that happens at the driver's position by decreasing  2.5$\%$ of the success rate. 

\subsection{Case Studies}
We totally tested 192 cases and results have been reported in the results section(III. A). To further display the performance of our method, we present the following 4 representative cases. In each case, we compare the performance of our method(APF+MPC) with HJ-reachability and also demonstrate its effectiveness in protecting the vehicle’s vulnerable region. As we mentioned in the algorithm framework part(II. E), we would increase the charge quantity at the driver's position when the protection mode is on. The selection of the charge quantity will be discussed in the discussion part(IV. B).
\subsubsection{Case I}
\begin{figure}[!ht]
\vspace*{-0.5cm}
\centering
\includegraphics[width=0.4\textwidth]{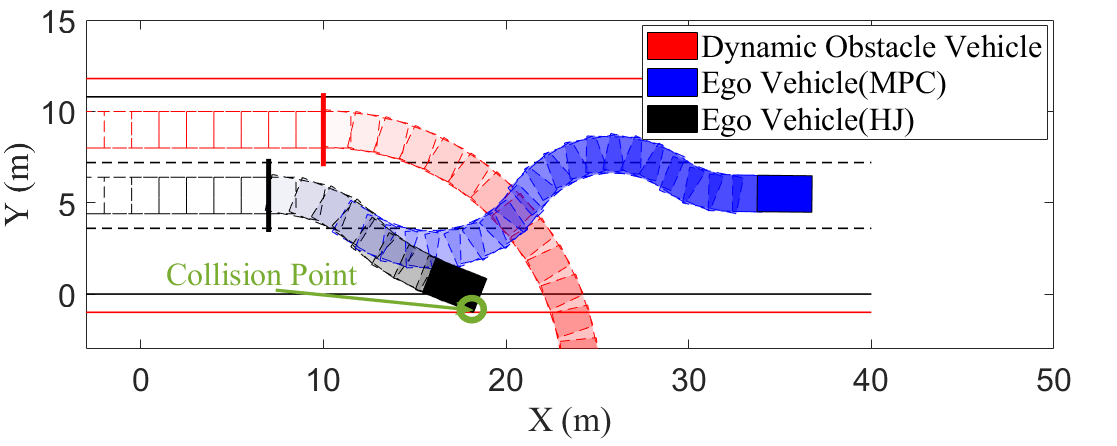}
\caption{Case I: Comparison between APF+MPC$\&$HJ-reachability}
\label{COMPHJ1}
\end{figure}

\begin{figure}[!ht]
\vspace*{-0.2cm}
\centering
\includegraphics[width=0.4\textwidth]{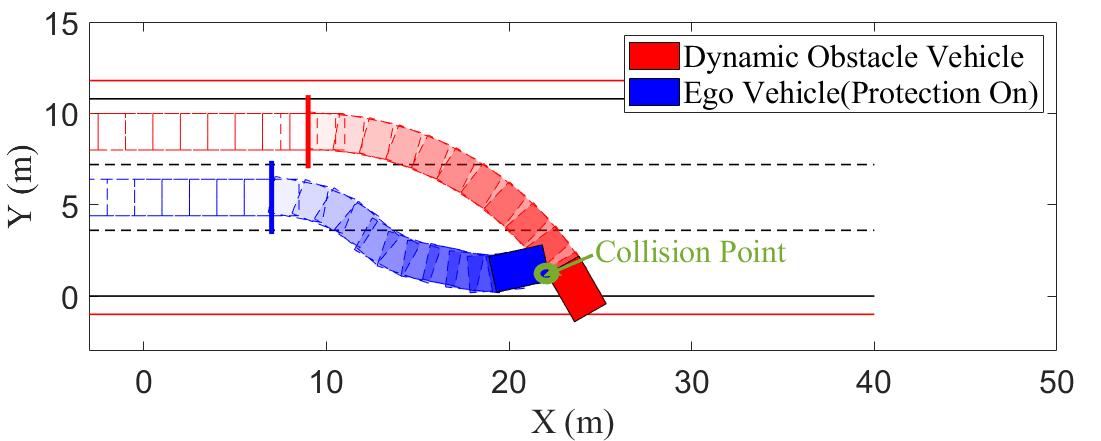}
\caption{Case I: Protection On}
\label{PTON1}
\end{figure}

\begin{figure}[!ht]
\vspace*{-0.2cm}
\centering
\includegraphics[width=0.4\textwidth]{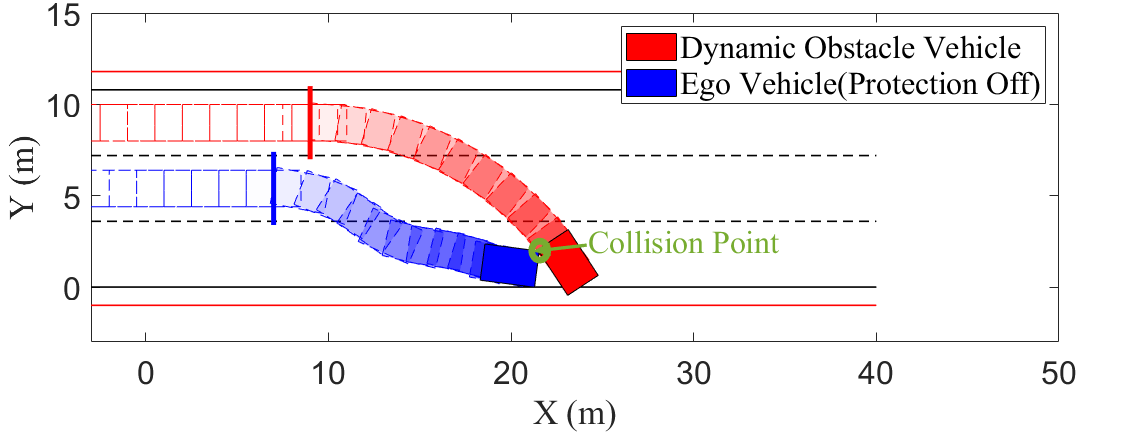}
\caption{Case I: Protection Off}
\label{PTOFF1}
\end{figure}

\begin{figure}[!ht]
\vspace*{-0.2cm}
\centering
\includegraphics[width=0.5\textwidth]{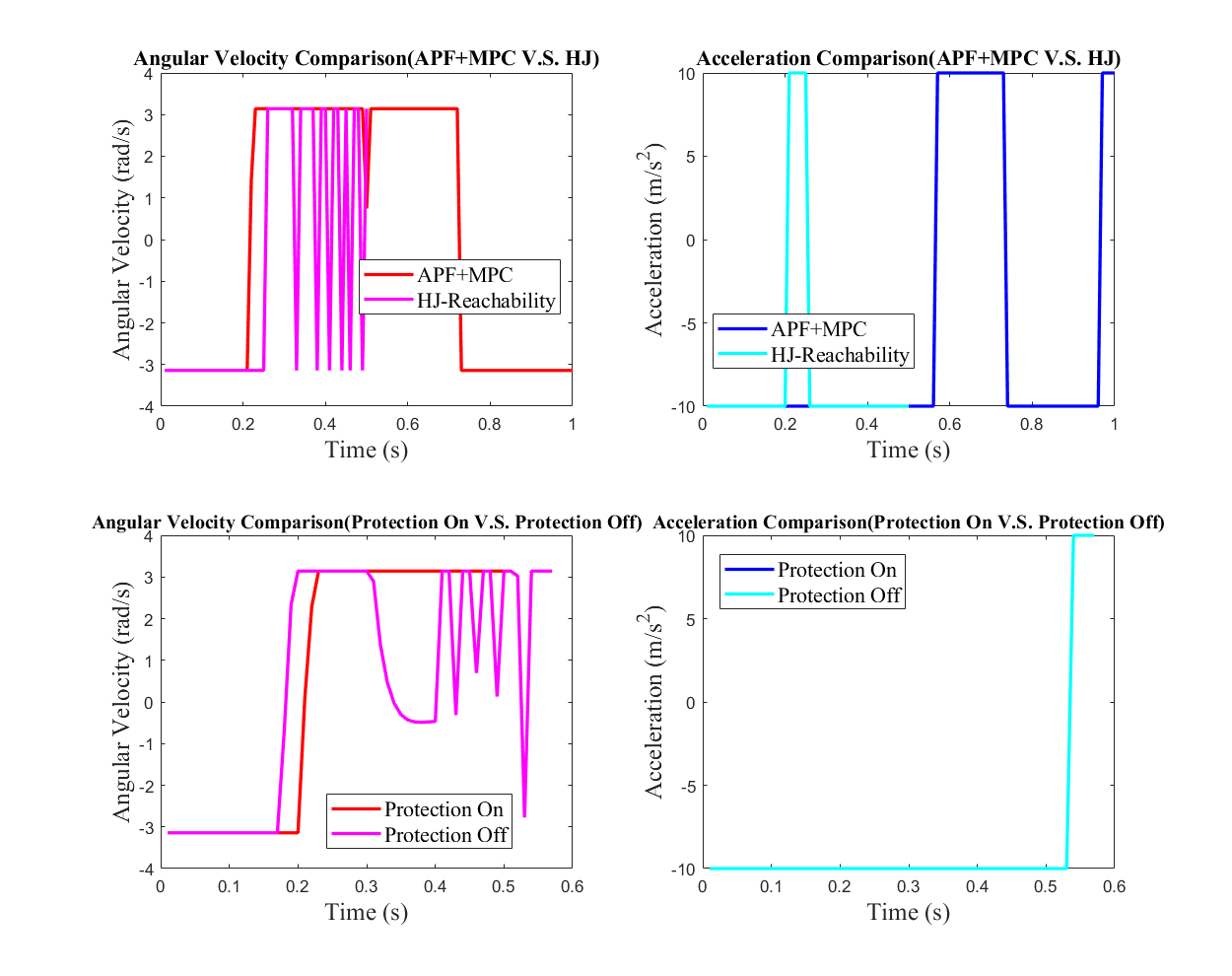}
\caption{Case I: Control input v.s. Time}
\label{CONINPUT1}
\end{figure}

Case I is from scenario 1 and it shows how the ego vehicle will react to a single obstacle vehicle in a highway scenario. In this case, the input of the obstacle vehicle is [-$\pi$/2; 3]. For the comparison between APF+MPC and HJ-reachability, the initial state of the ego vehicle is [7; 5.4; 0; 25] while the initial state of the obstacle vehicle is [10; 9; 0; 25]. For the comparison between protection on and protection off, the initial state of the ego vehicle is [7; 5.4; 0; 25] while the initial state of the obstacle vehicle is [9; 9; 0; 25]. Trajectories of vehicles are shown in Fig. \ref{COMPHJ1}, Fig.  \ref{PTON1}, and Fig.  \ref{PTOFF1}. The control inputs are shown in Fig.  \ref{CONINPUT1}.\\
\textbf{Comparison between APF+MPC$\&$HJ-reachability:} 
For the APF+MPC method, the ego vehicle will first try to turn right to get some space so that it could decelerate. Then it will turn left to avoid collision with the approaching vehicle. For the HJ-reachability method, the two separate tasks are avoiding collision with the obstacle vehicle and avoiding collision with roadsides. The ego vehicle will turn right in the beginning but it turns too much which makes it lose time to turn left and avoid the collision. It will keep switching between doing these two tasks and the input of the turning rate keeps bumping between the maximum value and the minimum value until the collision happens. \\
\textbf{Comparison between protection on$\&$protection off:} 
For both protection on and protection off mode, the ego vehicle will turn right and decelerate in the beginning. Then the ego vehicle will turn left and hit the obstacle vehicle. From the input plot Fig. \ref{CONINPUT1}, we could see that driver protection is realized by controlling the turning rate. From 0.3s to 0.6s, the ego vehicle will keep turning left when the protection mode is on rather than oscillating when the protection mode is off.      

\subsubsection{Case II}
\begin{figure}[!ht]
\centering
\includegraphics[width=0.4\textwidth]{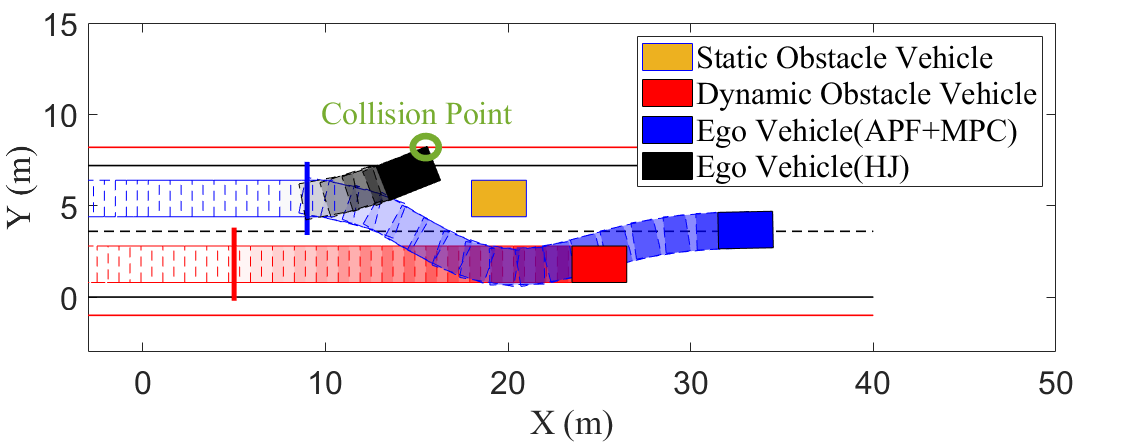}
\caption{Case II: Comparison between APF+MPC$\&$HJ-reachability}
\label{COMPHJ2}
\end{figure}

\begin{figure}[!htbp]
\centering
\includegraphics[width=0.4\textwidth]{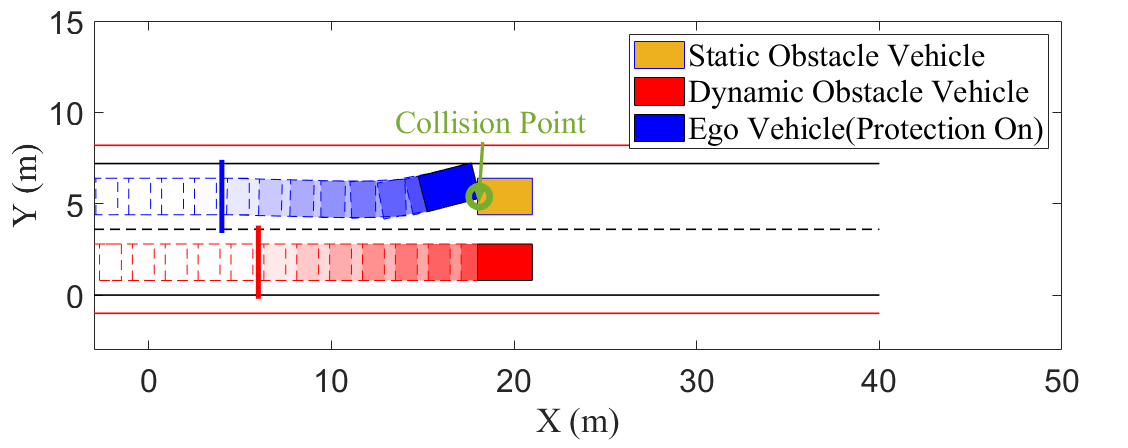}
\caption{Case II: Protection On}
\label{PTON2}
\end{figure}

\begin{figure}[!ht]
\vspace{-0.2cm}
\centering
\includegraphics[width=0.4\textwidth]{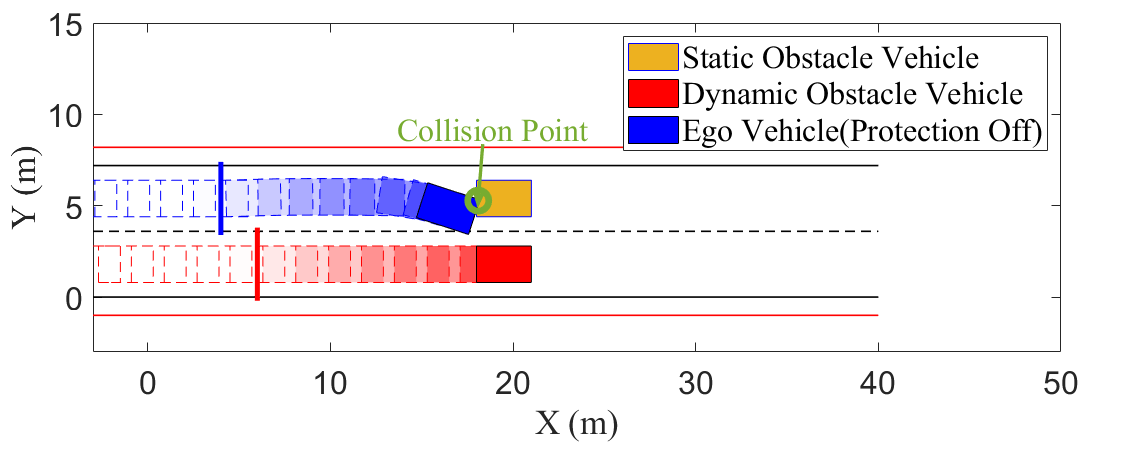}
\caption{Case II: Protection Off}
\label{PTOFF2}
\end{figure}

\begin{figure}[!ht]
\vspace{-0.2cm}
\centering
\includegraphics[width=0.5\textwidth]{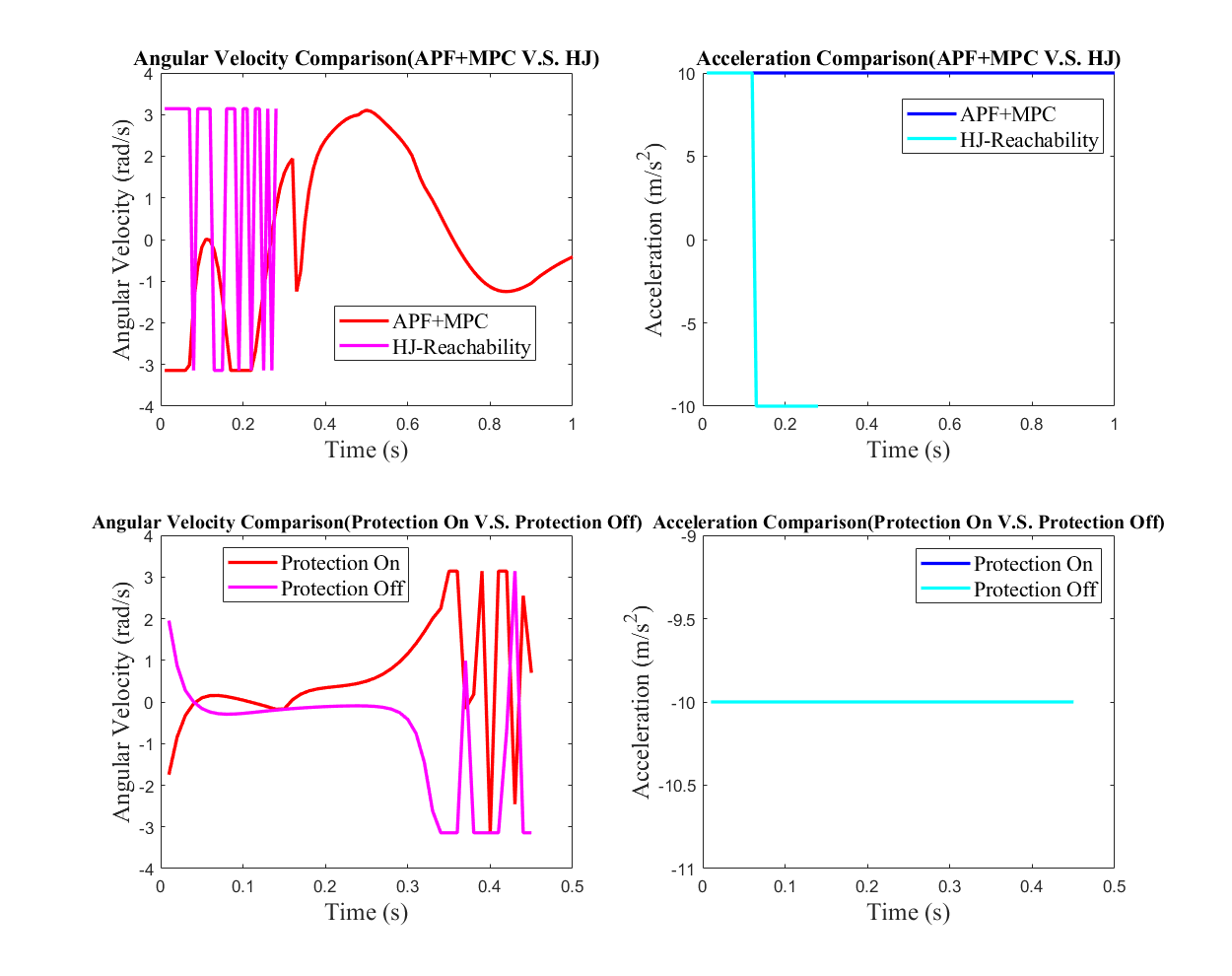}
\caption{Case II: Control input v.s. Time}
\label{CONINPUT2}
\end{figure}

Case II is from scenario 3 and it shows how the ego vehicle will react to two obstacle vehicles in a highway scenario. In this case, obstacle vehicle one will keep still and obstacle vehicle two will move at a constant speed. For the comparison between APF+MPC and HJ-reachability, the initial state of the ego vehicle is [9; 5.4; 0; 20] while the initial states for obstacle vehicle one and two are [19.5; 5.4; 0; 0] and [5; 1.8; 0; 20]. For the comparison between protection on and protection off, the initial state of the ego vehicle is [4; 5.4; 0; 30] while the initial states for obstacle vehicle one and two are [19.5; 5.4; 0; 0] and [6; 1.8; 0; 30]. Trajectories of vehicles are shown in Fig. \ref{COMPHJ2}, Fig. \ref{PTON2}, and Fig. \ref{PTOFF2}. The control inputs are shown in Fig. \ref{CONINPUT2}.\\
\textbf{Comparison between APF+MPC$\&$HJ-reachability:} 
For the APF+MPC method, the ego vehicle will keep accelerating to overtake the obstacle vehicle in the right lane. The ego vehicle will first turn right to avoid the collision with obstacle vehicle one and then turn left to avoid the collision with the roadside. For HJ-reachability, it has three tasks in this case which are avoiding collision with obstacle vehicle one, avoiding collision with obstacle vehicle two, and avoiding collision with the roadside. In the beginning, the priority for the ego vehicle is avoiding obstacle vehicle two in the right lane so it will turn left which makes it lose the opportunity to overtake obstacle vehicle two. The ego vehicle will then keep switching tasks from avoiding collision with obstacle vehicle one at the left lane and the roadside which makes the input of turning rate bump between the maximum value and the minimum value until the collision happens. \\
\textbf{Comparison between protection on$\&$protection off:} 
For both protection on and protection off mode, the ego vehicle will keep decelerating until it hits obstacle vehicle one in the left lane. From the input plot Fig. \ref{CONINPUT2}, we could see the mean difference shows up at 0.2s to 0.35s. During this time period, the ego vehicle will keep turning left when protection mode is on while the ego vehicle will keep turning right when protection mode is off. After 0.35s, the turning rate for both of them keeps bumping. The bumping is caused by the dominant term of the artificial potential function keeps changing.       
\subsubsection{Case III}
\begin{figure}[!ht]
\vspace{-0.3cm}
\centering
\includegraphics[width=0.4\textwidth]{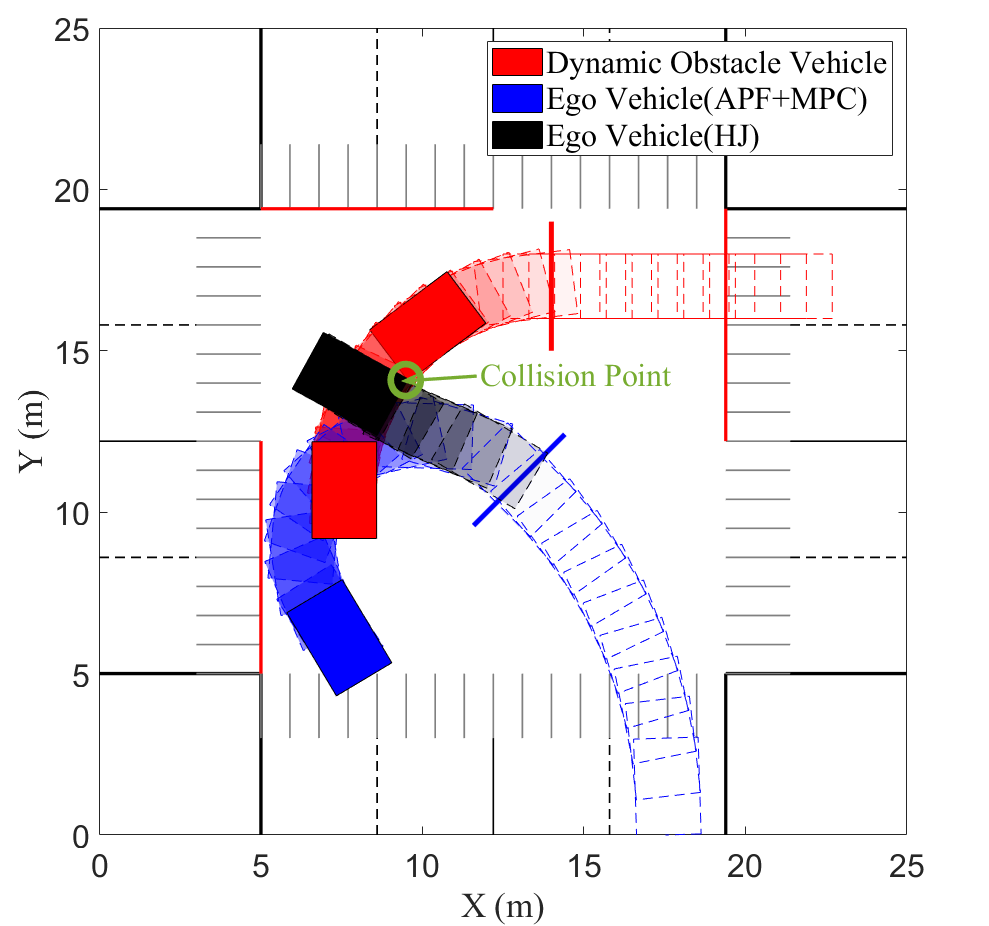}
\caption{Case III: Comparison between APF+MPC$\&$HJ-reachability}
\label{COMPHJ3}
\end{figure}

\begin{figure}[!ht]
\vspace{-0.3cm}
\centering
\includegraphics[width=0.4\textwidth]{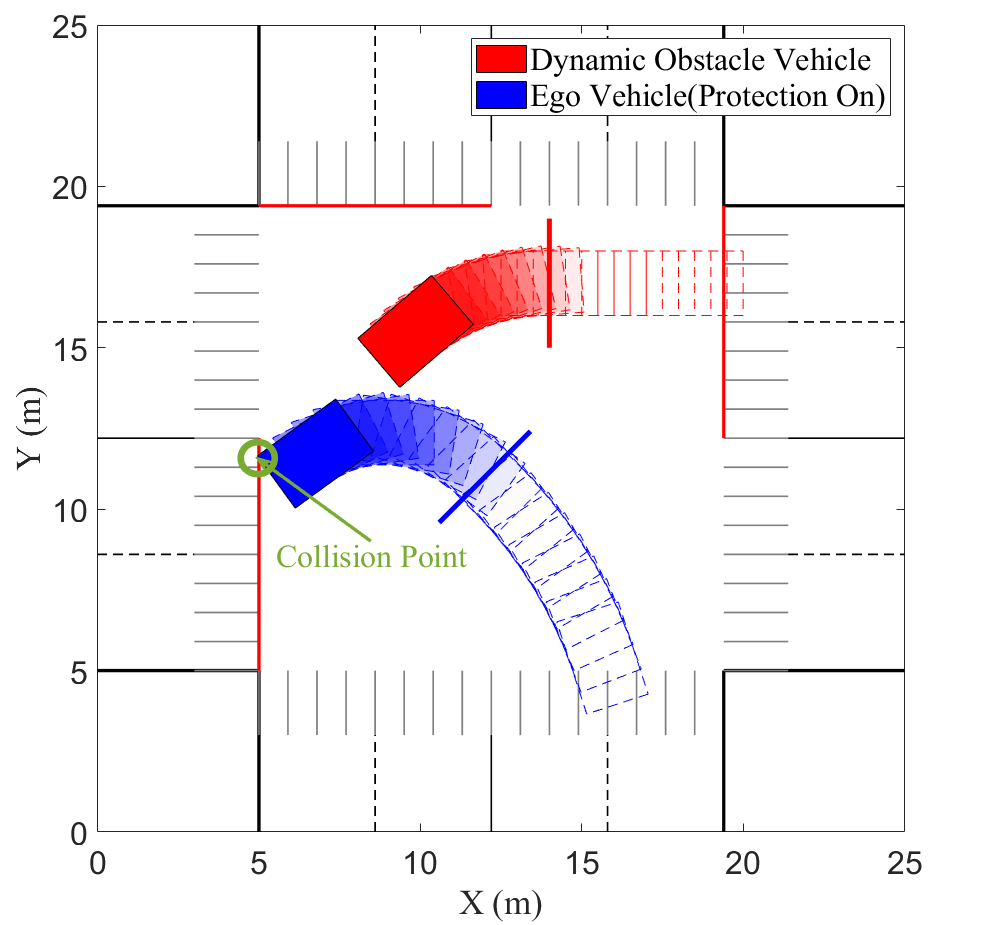}
\caption{Case III: Protection On}
\label{PTON3}
\end{figure}

\begin{figure}[!ht]
\centering
\includegraphics[width=0.4\textwidth]{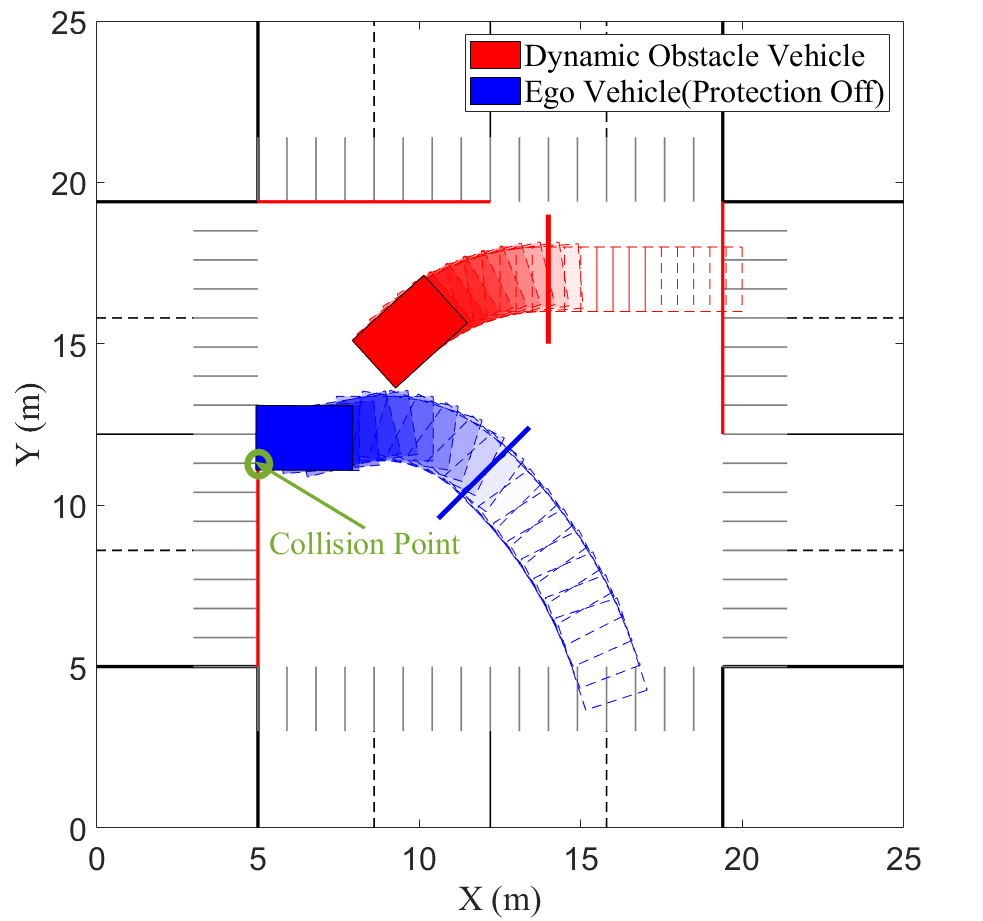}
\caption{Case III: Protection Off}
\label{PTOFF3}
\end{figure}

\begin{figure}[!ht]
\centering
\includegraphics[width=0.5\textwidth]{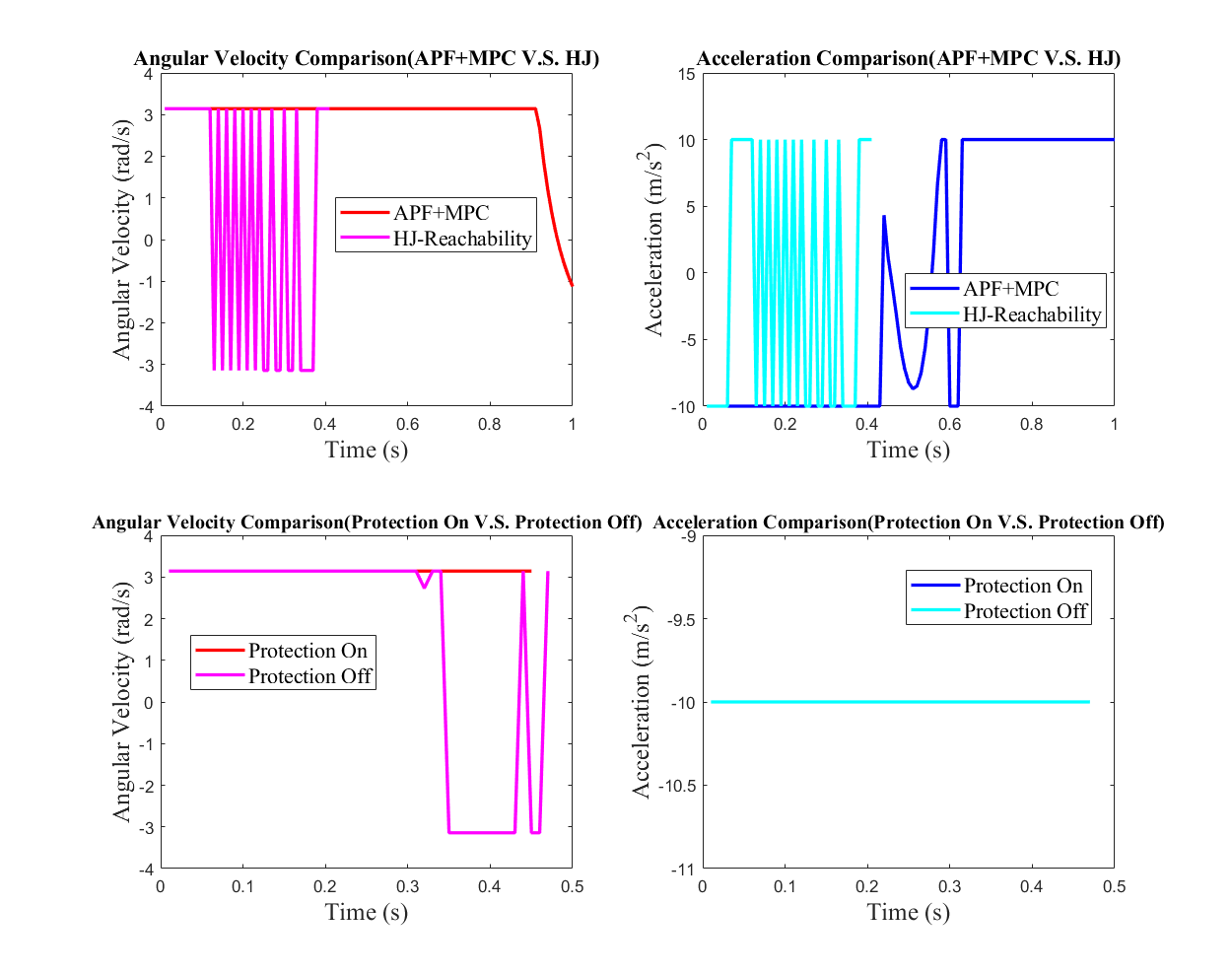}
\caption{Case III: Control input v.s. Time}
\label{CONINPUT3}
\end{figure}

Case III is a case from scenario 6 and it shows how the ego vehicle will react to a single obstacle vehicle in the intersection. In this case, the input of the obstacle vehicle is [$\pi$/2; 0]. For the comparison between APF+MPC and HJ-reachability, the initial state of the ego vehicle is [13; 11; 0.75$\pi$; 15] while the initial state of the obstacle vehicle is [14; 17; $\pi$; 10]. For the comparison between protection on and protection off, the initial state of the ego vehicle is [13; 11; 0.75$\pi$; 15] while the initial state of the obstacle vehicle is [12; 17; $\pi$; 10]. Trajectories of vehicles are shown in Fig. \ref{COMPHJ3}, Fig. \ref{PTON3}, and Fig. \ref{PTOFF3}. The control inputs are shown in Fig. \ref{CONINPUT3}.\\
\textbf{Comparison between APF+MPC$\&$HJ-reachability:} 
For the APF+MPC method, the ego vehicle will decelerate in the beginning to make a "U" turn and then accelerate to avoid the rear-end collision with the obstacle vehicle. The turning rate keeps at the maximum value which also proves the ego vehicle tends to make a "U" turn. For HJ-reachability, it has two tasks in this case which are avoiding collision with the obstacle vehicle and avoiding collision with the intersection roadside. In the beginning, the priority for the ego vehicle is avoiding the obstacle vehicle so it will turn left. Then the ego vehicle accelerates and tries to overtake the obstacle vehicle. However, the roadside limits the overtaking. The ego vehicle keeps switching tasks and the turning rate keeps bumping which is similar to case I. \\
\textbf{Comparison between protection on$\&$protection off:} 
For both protection on and protection off mode, the ego vehicle will keep decelerating until it hits the left avoidance area. From the input plot Fig. \ref{CONINPUT3}, we could see the mean difference shows up at 0.35s to 0.45s. During this time period, the ego vehicle will keep turning left when protection mode is on while the ego vehicle will keep turning right when protection mode is off.

\subsubsection{Case IV}
\begin{figure}[!ht]
\centering
\includegraphics[width=0.4\textwidth]{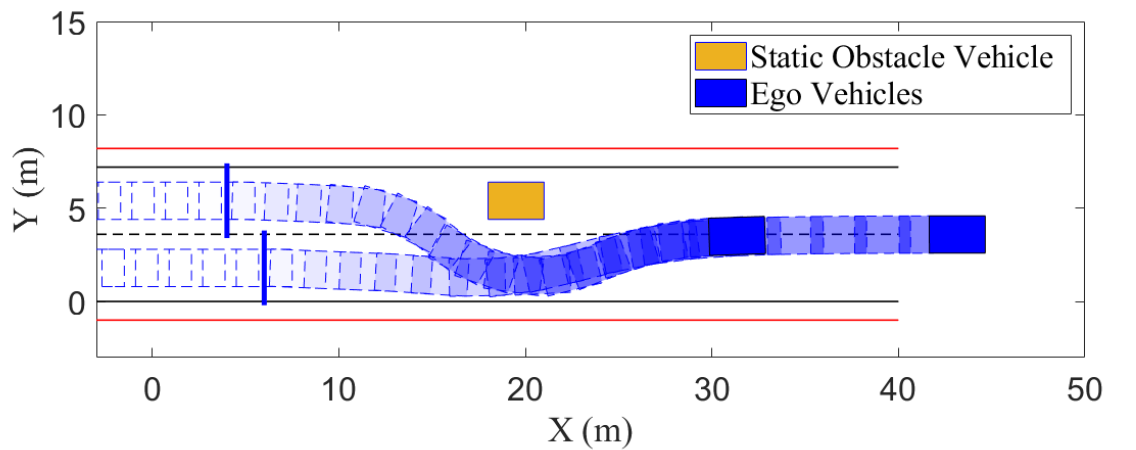}
\caption{Case IV: Multi-vehicle cooperative control}
\label{MVCC}
\end{figure}

\begin{figure}[!ht]
\centering
\includegraphics[width=0.4\textwidth]{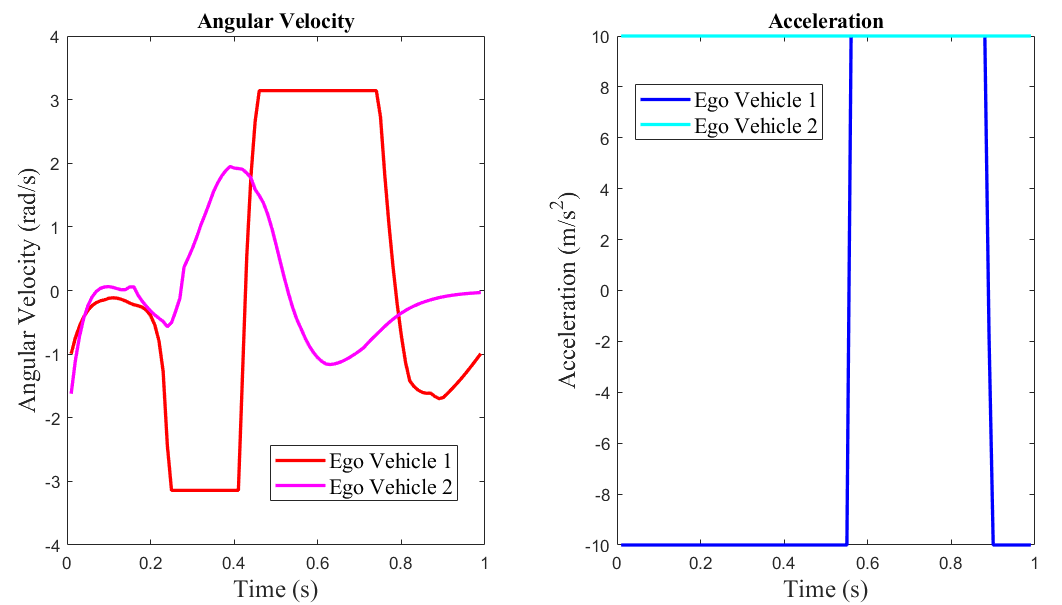}
\caption{Case IV: Control input v.s. Time}
\label{CONINPUT4}
\end{figure}

The parameter settings for Case IV are the same as the comparison between protection on$\&$protection off in Case II. The difference is we assume we could also control obstacle vehicle two in Case II. By doing cooperative control, we could avoid an inevitable collision if only controlling a single vehicle.
\indentpar
We define the ego vehicle in the left lane as ego vehicle one and the ego vehicle in the right lane as ego vehicle two. As we could see from Fig. \ref{MVCC} and Fig. \ref{CONINPUT4}, in the beginning, ego vehicle two will turn right and keep accelerating to give more space for ego vehicle one. Ego vehicle one will turn right and decelerate to avoid collision with the obstacle vehicle. Then it will accelerate to move away from the obstacle vehicle until it gets close to the ego vehicle two. 

\section{Discussion}
\subsection{HJ-reachability Revisit}
When decomposing the whole system into several subsystems, the problem of sharing control will show up ~\cite{17Chen}. For each individual subsystem, the control input will always try to make the value function as large as possible which means moving away from the avoid set. Recall that in the algorithm framework part(II. F), we define $V_1$ and $V_2$ as value functions of two subsystems. The control input that makes $V_1$ larger may decrease $V_2$. When one of them becomes smaller than zero, a collision will happen. That's the main reason that HJ-reachability will have a lower success rate compared with our method. 
\indentpar
Another problem with HJ-reachability is it doesn't consider collision mitigation when the collision is inevitable. We think this problem could be solved by combining the artificial potential function we proposed with HJ-reachability and we explain it in future work part. 
\subsection{Parameters Selection}
\begin{figure}[!ht]
\centering
\includegraphics[width=0.4\textwidth]{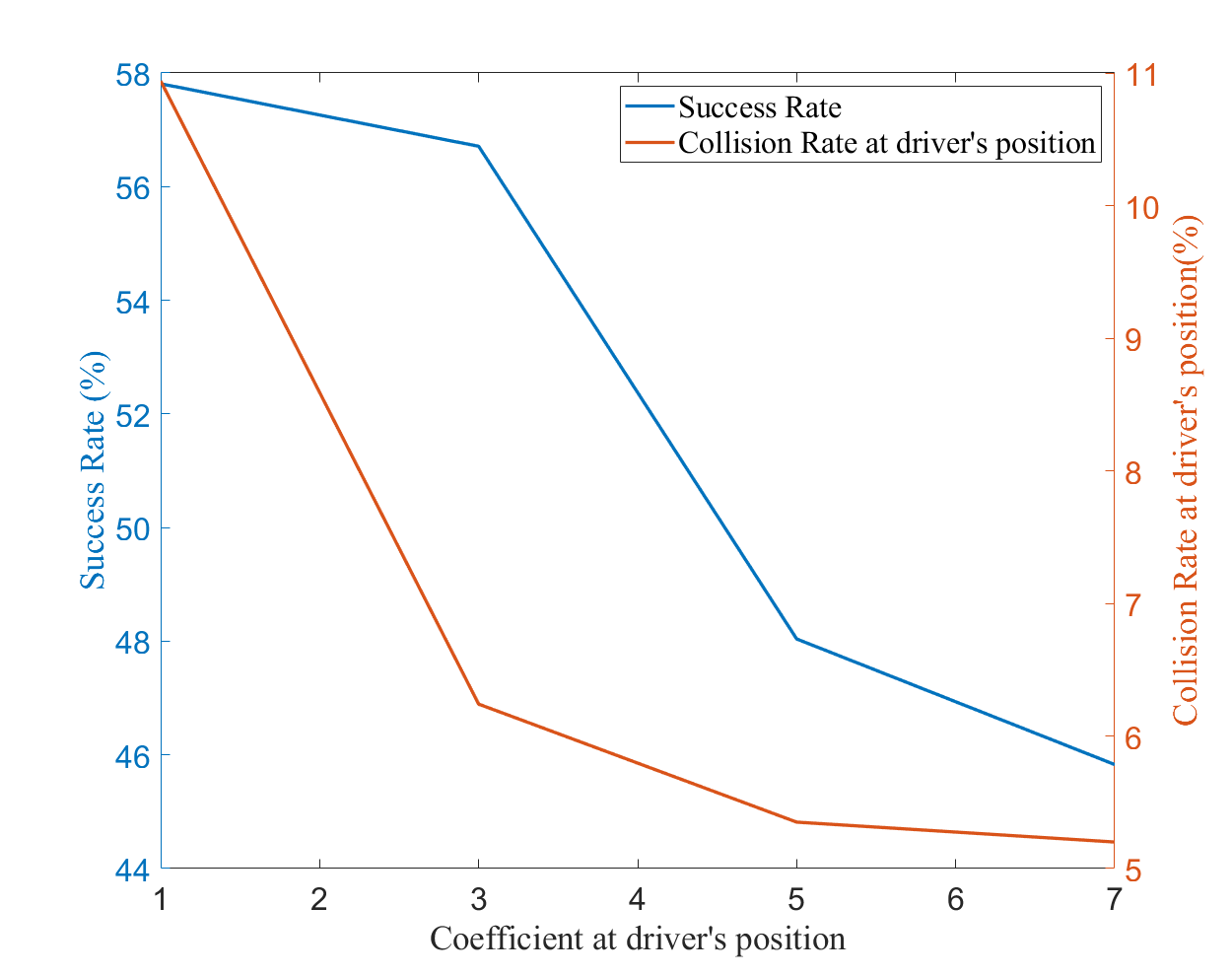}
\caption{Successful rate and collision rate at driver's position for different driver's position's charge quantities}
\label{SRDC}
\end{figure}

There are two kinds of parameters in our artificial potential function. One is weights between different obstacles and the other is charge quantities for the additional point charge in Fig. \ref{PE}. 
\indentpar
The weights for different obstacles will affect the location of the local minimum point in the potential field and thus affect the success rate and the priority for avoiding which obstacle. For the local minimum problem, the use of MPC in our method could partially solve it. The performance will become better as the increasing of prediction horizon while the computational time will also increase. We use the same weight and the same line density $\lambda$ for all obstacles since we don't assign priority to different obstacles and we want the same length in different obstacles will have the same effect for the ego vehicle. 
\indentpar
The charge quantity for the additional point charge will make a trade-off between the success rate and the rate of collision at the driver's position. If we increase the charge quantity of the point at the driver's position to an extremely large number, the model will degrade to a point rather than taking the shape of the ego vehicle into consideration. That will avoid most of the collisions happening at the driver's place but decrease the success rate significantly. To efficiently choose a suboptimal charge quantity in the driver's place, we test 4 different charge quantities at the driver's position which are 1C, 3C, 5C, and 7C. As we could see from Fig. \ref{SRDC},  3C is a relatively better choice which has a much lower collision rate at the driver's position compared with 1C and a much higher success rate compared with 5C and 7C. Thus, we choose to use 3C at the driver's position when the protection is on.
\indentpar
We also compare the results of our parameter selection with the results from the genetic method with the proposed artificial potential function. The genetic algorithm is inspired by the nature of biological evolution and it has been used for deciding the coefficient of artificial potential function(GA+APF) \cite{25Vadakkepat, 26Xu}. The genetic algorithm could also deal with multi-objective optimization so that other factors of the path could be taken into consideration such as smoothness and length. In our case, two objective functions are the maximum potential energy of the whole ego vehicle and the maximum potential energy of the driver's position along the path. We hope to minimize both of them to balance the trade-off problem. However, the algorithm is also required to be implemented in real-time and there isn't enough time to wait for the genetic algorithm to converge. We run the genetic algorithm for 60s in each case to see its performance. The actual time that could be given in a real scenario should be much less than 60s. As we could see from Table \ref{tab3}, compared with APF+MPC(protection on), the collision rate at the driver's position decreases from 6.24$\%$ to 1.04$\%$ but the success rate also decreases from 56.71$\%$ to 48.95$\%$. That shows GA+APF has better driver protection performance but decreases the success rate. Thus, parameter optimization methods run online could not give a Pareto improvement in a limited time.   

\begin{table}
\begin{center}
\caption{Successful rate and collision rate at driver's position(GA+APF)}
\label{tab3}
\begin{tabular}{| c | c | c |}
\hline
Scenario & Success Rate & Collision rate at driver's position  \\
\hline
1 & 54.17$\%$ & 0$\%$ \\
\hline 
2 & 29.17$\%$ & 0$\%$ \\
\hline
3 & 50$\%$ & 0$\%$ \\
\hline 
4 & 37.5$\%$ & 4.17$\%$ \\
\hline
5 & 87.5$\%$ & 0$\%$ \\
\hline
6 & 29.17$\%$ & 0$\%$ \\
\hline 
7 & 58.33$\%$ & 4.17$\%$ \\
\hline
8 & 45.83$\%$ & 0$\%$ \\
\hline
\end{tabular}
\end{center}
\end{table}
\section{Conclusion}
In this paper, we build a safe controller by combining the model predictive control method and artificial potential function for emergency collision avoidance. A new artificial potential function is proposed which is inspired by line charge. There are two main advantages of using this new artificial potential function: 1. It considers the shape of obstacle vehicles, road structures, and the ego vehicle which could be used in more road structures and provides more feasible solutions in an emergency; 2. Specific part collision mitigation could be realized by turning the parameters of the artificial potential function. 
\indentpar
Simulation results in 192 different cases from 8 different scenarios show that our approach works well in different complex scenarios and could be used in multi-vehicle cooperative control. Compared with results from HJ-reachability in the unicycle model, the success rate of our approach is 20$\%$ higher. Our approach could also decrease 43$\%$ of collision in the driver's position with a 2$\%$ decrease in the success rate. The method is further validated in a dynamic bicycle model. For the parameters selection problem, we show that the online parameter optimization method(genetic algorithm) which runs for a limited time could not give a Pareto improvement for success rate and collision rate in the driver's position.  
\indentpar
Although the present work demonstrates the feasibility and effectiveness of the proposed MPC+AFC method and compares its advantages over reachability sets, there are still opportunities to extend this research further. We highlight what can be done for improvements for the benefit of researchers in this field. They are summarised as follows. First, one could combine the proposed artificial potential function with HJ-reachability. A control barrier function could be constructed using the proposed artificial potential function \cite{24Singletary}, and we could use it as the value function in HJ-reachability. Collision mitigation would be considered in the backward reachable set calculation using the new value function. Second, reinforcement learning has been used for training the controller based on the classical artificial potential function \cite{27Yao}. New research can use a similar method to train our controller offline, based on the artificial potential function we proposed, to realize better performance in driver protection and success rate. Third, one can use real traffic data to simulate the traffic flow and take the estimation uncertainties \cite{60Ghorai}, perception uncertainties \cite{59Sridhar} and prediction uncertainties \cite{28Nayak} into consideration. This approach allows for building a heat map with the proposed artificial potential function and the real traffic flow data. Combining the heat map with a probability map, both the density and the randomness of the traffic flow could be considered. These are practical future research ideas that can be useful extensions of the present work.
{\appendices
\section{Basic knowledge of HJ-reachability~\cite{55mitchell}}\label{HJIntro}
Hamilton-Jacobi reachability computes the backwards reachable set of a system. The backwards reachable set contains states from which trajectories start that will reach the target set as shown in Fig. \ref{HJF}. 
\begin{figure}[!ht]
\centering
\includegraphics[width=0.4\textwidth]{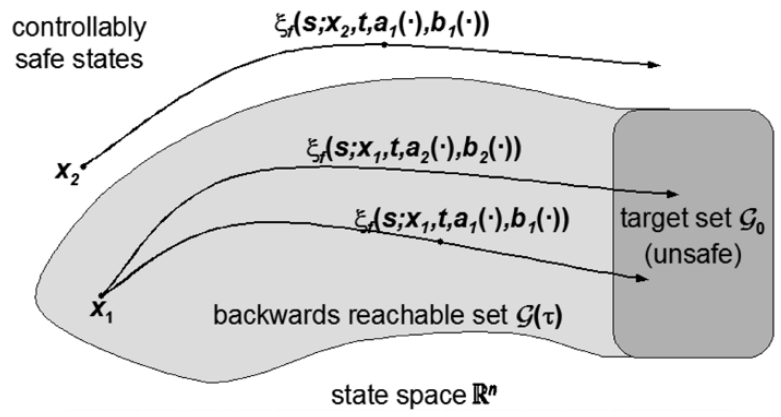}
\caption{Target Set and backward reachable. Figure taken from \cite{55mitchell}}
\label{HJF}
\end{figure}
In our case, the target set represents states where collisions have already happened and the backwards reachable set represents states that will lead to a collision. Thus, the system should avoid entering the backwards reachable set during path planning. 
\indentpar
Assumed that there are only an ego vehicle and an obstacle vehicle in the environment. The ego vehicle will try to avoid entering the target set while the obstacle vehicle tries to steer the system toward the target set. The system dynamics could be represented as: 
\begin{equation}\label{HJSYSDY}
    \dot{x} = f(x, a, b)
\end{equation}
where $x$ is the states of the system, $a(t)$ is the input of the ego vehicle and $b(t)$ is the input of obstacle vehicle. Trajectories of the system are solutions of (\ref{HJSYSDY}) which is: 
\[
\xi_f(s;x,t,a,b): [t, 0] \rightarrow R^n.
\]
They need to satisfy the initial condition $\xi_f(t;x,t,a,b) = x_0$ and the differential equation that is: 
\[
\frac{d}{ds} \xi_f(s;x,t,a,b) = f(\xi_f(s;x,t,a,b),a,b).
\]
\indentpar
The target set $\mathcal{G}_0 $ is defined as the zero sublevel set of a continuous function $g$: $\mathbb{R}^n \rightarrow \mathbb{R}$
\[
\mathcal{G}_0 = \{ x \in \mathbb{R}^n \vert g(x) \le 0 \}.
\]
The backwards reachable set is represented as:
\begin{equation}
\begin{split}
    \mathcal{G}(t) = \{x: \exists \gamma \in \Gamma(t), \forall a(\cdot) \in \mathbb{A}, \\
    \xi(0;x,t,a,\gamma[a](\cdot)) \in \mathcal{G}_0\}
\end{split}
\end{equation}
where $\mathbb{A}$ is the feasible set of inputs for ego vehicle and $\Gamma(\cdot)$ represents the feasible set of strategies for the obstacle vehicle.
\indentpar
Let $v$ be the viscosity solution of the HJI equation
\[
D_t v(x,t) + min[0, H(x, D_x v(x,t))] = 0, v(x, 0) = g(x)
\]
where $H$ is called the Hamiltonian and is given by
\[
H(x, p) = \underset{a\in A}{\mathtt{max}} \ \underset{b\in B}{\mathtt{min}} \ p^T f(x, a, b).
\]
Then, the backward reachable set could be represented as: 
\[
\mathcal{G}_t = \{ x\in \mathbb{R}^n \vert v(x,t) \le 0\}.
\]
and the optimal control for the ego vehicle is: 
\[
a^*(t,x) = \mathtt{arg} \ \underset{a\in A}{\mathtt{max}} \ \underset{b\in B}{\mathtt{min}} \ p^T f(x, a, b),
\]
which will minimize the signed distance between the ego vehicle and the target set at the end of the time period. If $x(t) \in \mathcal{G}_t^C $, the ego vehicle has a control sequence that will keep it out of the target set, irrespective of the control of the obstacle vehicle.}

\bibliographystyle{IEEEtran}
\bibliography{99_refs}

\vspace{11pt}
\begin{IEEEbiography}
[{\includegraphics[width=1in,height=1.25in,clip,keepaspectratio]{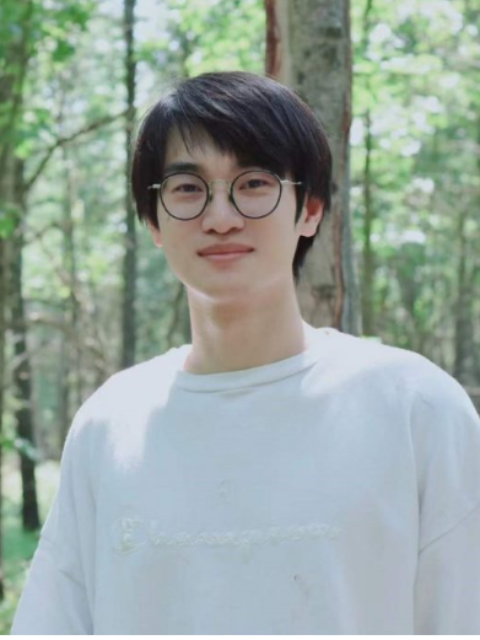}}]{Xu Shang} received the B.S. degree in mechanical engineering from Shanghai Jiao Tong university and the M.S. degree from University of Michigan. He is currently pursuing
the Ph.D. degree in mechanical engineering from
Virginia Tech. His current research interests include robotics
and control, decision making, path planning, and connected vehicles.
\end{IEEEbiography}
\vspace{11pt}
\begin{IEEEbiography}[{\includegraphics[width=1in,height=1.25in,clip,keepaspectratio]{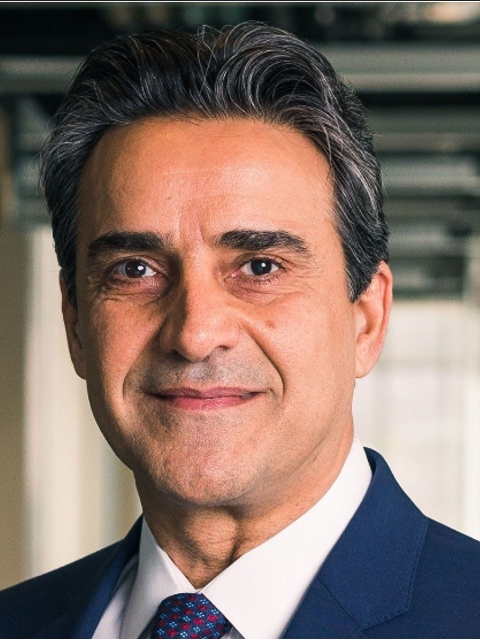}}]{Azim Eskandarian} received the B.S. and D.Sc. degrees from George
Washington University (GWU) and the M.S. degree
from Virginia Tech all in mechanical engineering.
He was a Professor of engineering and applied
science with GWU and the Founding Director of
the Center for Intelligent Systems Research from
1996 to 2015. From 2002 to 2015, he was the
Director of the Transportation Safety and Security
University Area of Excellence. In 1992, he was
a Co-Founder of the National Crash Analysis
Center. From 1998 to 2002 and 2013 to 2015, he was the Director of
the National Crash Analysis Center. He was an Assistant Professor with
Pennsylvania State University, York, PA, USA, from 1989 to 1992, and an
Engineer/Project Manager in industry from 1983 to 1989. He has been a
Professor and the Head of the Mechanical Engineering Department with
Virginia Tech (VT), since 2015. In 2018, he became the Nicholas and
Rebecca Des Champs Chaired Professor. He established the Autonomous
Systems and Intelligent Machines Laboratory at VT to conduct research on
intelligent and autonomous vehicles and mobile robots. He is a fellow of
ASME and a member of SAE professional societies. He received the IEEE
ITS Society’s Outstanding Researcher Award in 2017 and GWU’s School of
Engineering Outstanding Researcher Award in 2013.
\end{IEEEbiography}
\vfill
\end{document}